\pdfoutput=1

\documentclass[11pt]{article}

\usepackage[]{acl}

\usepackage{times}
\usepackage{latexsym}

\usepackage[T1]{fontenc}

\usepackage[utf8]{inputenc}

\usepackage{microtype}

\usepackage{inconsolata}

\usepackage{graphicx}
\usepackage{booktabs}  
\usepackage{multirow}  
\usepackage{colortbl}  
\usepackage{tabularx}  
\usepackage{makecell}  
\usepackage{xltabular} 
\usepackage{amsmath}   
\usepackage{subcaption} 

\usepackage{pgfplots}  
\pgfplotsset{compat=newest} 
\usepackage{soul}

\usepackage{xspace}

\definecolor{shadow}{gray}{.9} 
\definecolor{cGray}{RGB}{186,186,186}
\definecolor{cBlue}{RGB}{167,192,223}
\definecolor{Mizuiro}{RGB}{4,178,217}
\definecolor{cGreen}{RGB}{160,208,208}
\definecolor{Summer}{RGB}{84,217,159}
\definecolor{cRed}{RGB}{240,128,128}
\definecolor{cYellow}{RGB}{255,170,0}

\title{Revealing the Parallel Multilingual Learning within Large Language Models}

\author{Yongyu Mu\textsuperscript{1}\thanks{\xspace\xspace Equal contribution.}, Peinan Feng\textsuperscript{1}\footnotemark[1], Zhiquan Cao\textsuperscript{1}\footnotemark[1], Yuzhang Wu\textsuperscript{1}, Bei Li\textsuperscript{1}, Chenglong Wang\textsuperscript{1}, \\
{\bf Tong Xiao\textsuperscript{1,2}\thanks{\xspace\xspace Corresponding author.}, Kai Song\textsuperscript{3}, Tongran Liu\textsuperscript{4}, Chunliang Zhang\textsuperscript{1,2} \and Jingbo Zhu\textsuperscript{1,2}} \\
	\textsuperscript{1}NLP Lab, School of Computer Science and Engineering, Northeastern University, Shenyang, China\\
	\textsuperscript{2}NiuTrans Research, Shenyang, China\\
        \textsuperscript{3}Bytedance, Seattle\\
        \textsuperscript{4}CAS Key Laboratory of Behavioral Science, Institute of Psychology, CAS, Beijing, China\\
	\ttfamily{lixiaoyumu9@gmail.com} \ttfamily{\{xiaotong,zhujingbo\}@mail.neu.edu.cn}
}

\begin{document}
\maketitle

\begin{abstract}
Large language models (LLMs) can handle multilingual and cross-lingual text within a single input; however, previous works leveraging multilingualism in LLMs primarily focus on using English as the pivot language to enhance language understanding and reasoning. Given that multiple languages are a compensation for the losses caused by a single language's limitations, it's a natural next step to enrich the model’s learning context through the integration of the original input with its multiple translations. In this paper, we start by revealing that LLMs learn from \textbf{P}arallel \textbf{M}ultilingual \textbf{I}nput (\textbf{PMI}). Our comprehensive evaluation shows that PMI enhances the model's comprehension of the input, achieving superior performance than conventional in-context learning (ICL). Furthermore, to explore how multilingual processing affects prediction, we examine the activated neurons in LLMs. Surprisingly, involving more languages in the input activates fewer neurons, leading to more focused and effective neural activation patterns. This neural reaction coincidentally mirrors the neuroscience insight about synaptic pruning, highlighting a similarity between artificial and biological `brains'. Our parallel multilingual data and code could be found at \url{https://github.com/takagi97/LLMs-are-parallel-multilingual-learners}.
\end{abstract}


\section{Introduction}
Many of the recent large language models (LLMs) are multilingual. Unlike language-specific NLP systems, such as machine translation systems specialized to a given language pair, these models are generally trained on large-scale multilingual datasets, using a unified vocabulary. Because of this training approach, it is possible to learn a universal representation of texts across different languages. Therefore, the resulting models can be directly applied to a variety of multilingual and cross-lingual tasks. For example, most commercialized LLMs can respond to user queries in different languages, without needing to specify what languages are used. More recently, the multilingual capabilities of these models have been shown to help cross-lingual in-context learning (ICL). By providing simple prompts involving cross-lingual thinking and reasoning, LLMs can understand and generate text in languages that were less represented in the training data \cite{DBLP:conf/emnlp/0001CWHC23,DBLP:conf/emnlp/HuangTZZSXW23,DBLP:journals/corr/abs-2311-08711,DBLP:journals/corr/abs-2306-11372}.

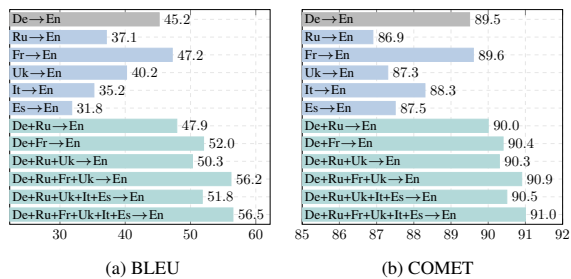
\begin{figure}
\centering
\begin{tikzpicture} [scale=0.5]

\begin{axis} [
    at={(0,0)},
    xbar, 
    xmin=28,
    ytick={1,2,3,4,5,6,7,8,9,10,11,12},
    y tick style={draw=none},
    yticklabels={\empty},
    yticklabel style={
        align=left,
    },
    nodes near coords,
    nodes near coords style={
        font=\normalsize,
        /pgf/number format/.cd,
        fixed zerofill,
        precision=1,
    },
    bar shift=0pt,
    bar width=10pt,
    xmajorgrids,
    ymajorgrids,
    grid style={draw=black!10, dashed},
    enlarge x limits=0.2,
    enlarge y limits=0.05,
]

\addplot [
    draw = cGray,
    fill = cGray,
]  coordinates {(45.2,12)};
\addplot [
    draw = cBlue!80,
    fill = cBlue!80,
]  coordinates {(37.1,11)(47.2,10)(40.2,9)(35.2,8)(31.8,7)};
\addplot [
    draw = cGreen!80,
    fill = cGreen!80
]  coordinates {(47.9,6)(52.0,5)(50.3,4)(56.2,3)(51.8,2)(56.5,1)};

\end{axis} 
\node (12)[anchor=west,font=\fontsize{5}{5}\selectfont] at(-0.2,5.44){De$\rightarrow$En};
\node (11)[anchor=west,font=\fontsize{5}{5}\selectfont] at(-0.2,4.97){Ru$\rightarrow$En};
\node (10)[anchor=west,font=\fontsize{5}{5}\selectfont] at(-0.2,4.50){Fr$\rightarrow$En};
\node (9)[anchor=west,font=\fontsize{5}{5}\selectfont] at(-0.2,4.03){Uk$\rightarrow$En};
\node (8)[anchor=west,font=\fontsize{5}{5}\selectfont] at(-0.2,3.56){It$\rightarrow$En};
\node (7)[anchor=west,font=\fontsize{5}{5}\selectfont] at(-0.2,3.09){Es$\rightarrow$En};
\node (6)[anchor=west,font=\fontsize{5}{5}\selectfont] at(-0.2,2.62){De+Ru$\rightarrow$En};
\node (5)[anchor=west,font=\fontsize{5}{5}\selectfont] at(-0.2,2.15){De+Fr$\rightarrow$En};
\node (4)[anchor=west,font=\fontsize{5}{5}\selectfont] at(-0.2,1.68){De+Ru+Uk$\rightarrow$En};
\node (3)[anchor=west,font=\fontsize{5}{5}\selectfont] at(-0.2,1.21){De+Ru+Fr+Uk$\rightarrow$En};
\node (2)[anchor=west,font=\fontsize{5}{5}\selectfont] at(-0.2,0.74){De+Ru+Uk+It+Es$\rightarrow$En};
\node (1)[anchor=west,font=\fontsize{5}{5}\selectfont] at(-0.2,0.27){De+Ru+Fr+Uk+It+Es$\rightarrow$En};

\node (titlea)[anchor=center,font=\scriptsize] at(3.5,-1.2){(a) BLEU};

\begin{axis} [
    at={(20em,0)},
    xbar, 
    xmin=86,
    ytick={1,2,3,4,5,6,7,8,9,10,11,12},
    y tick style={draw=none},
    yticklabels={\empty},
    yticklabel style={align=center},
    nodes near coords,
    nodes near coords style={
        font=\normalsize,
        /pgf/number format/.cd,
        fixed zerofill,
        precision=1,
    },
    bar shift=0pt,
    bar width=10pt,
    xmajorgrids,
    ymajorgrids,
    grid style={draw=black!10, dashed},
    enlarge x limits=0.2,
    enlarge y limits=0.05,
]

\addplot [
    draw = cGray,
    fill = cGray,
]  coordinates {(89.5,12)};
\addplot [
    draw = cBlue!80,
    fill = cBlue!80,
]  coordinates {(86.9,11)(89.6,10)(87.3,9)(88.3,8)(87.5,7)};
\addplot [
    draw = cGreen!80,
    fill = cGreen!80,
]  coordinates {(90.0,6)(90.4,5)(90.3,4)(90.9,3)(90.5,2)(91.0,1)};

\end{axis} 

\node (12)[anchor=west,font=\fontsize{5}{5}\selectfont] at(7.5,5.44){De$\rightarrow$En};
\node (11)[anchor=west,font=\fontsize{5}{5}\selectfont] at(7.5,4.97){Ru$\rightarrow$En};
\node (10)[anchor=west,font=\fontsize{5}{5}\selectfont] at(7.5,4.50){Fr$\rightarrow$En};
\node (9)[anchor=west,font=\fontsize{5}{5}\selectfont] at(7.5,4.03){Uk$\rightarrow$En};
\node (8)[anchor=west,font=\fontsize{5}{5}\selectfont] at(7.5,3.56){It$\rightarrow$En};
\node (7)[anchor=west,font=\fontsize{5}{5}\selectfont] at(7.5,3.09){Es$\rightarrow$En};
\node (6)[anchor=west,font=\fontsize{5}{5}\selectfont] at(7.5,2.62){De+Ru$\rightarrow$En};
\node (5)[anchor=west,font=\fontsize{5}{5}\selectfont] at(7.5,2.15){De+Fr$\rightarrow$En};
\node (4)[anchor=west,font=\fontsize{5}{5}\selectfont] at(7.5,1.68){De+Ru+Uk$\rightarrow$En};
\node (3)[anchor=west,font=\fontsize{5}{5}\selectfont] at(7.5,1.21){De+Ru+Fr+Uk$\rightarrow$En};
\node (2)[anchor=west,font=\fontsize{5}{5}\selectfont] at(7.5,0.74){De+Ru+Uk+It+Es$\rightarrow$En};
\node (1)[anchor=west,font=\fontsize{5}{5}\selectfont] at(7.5,0.27){De+Ru+Fr+Uk+It+Es$\rightarrow$En};

\node (titleb)[anchor=center,font=\scriptsize] at(11,-1.2){(b) COMET};

\end{tikzpicture}
\vspace{-1.8em}
\caption{Comparing the effectiveness of our \colorbox{cGreen}{PMI} versus \colorbox{cGray}{direct} and \colorbox{cBlue}{pivot} translation on the Qwen-14B model and the FLORES-200 dataset. We also provide the results of ChatGPT in Table \ref{tab:t1}.}
\label{fig:f1}
\end{figure}

Despite the apparent usefulness of multilingualism in LLMs, previous work has primarily focused on using English as the pivot language in language understanding and reasoning. It is a natural next step to incorporate more languages and investigate how these languages are simultaneously processed in LLMs. In this paper, we explore methods that make use of parallel multilingual input (PMI) in ICL and explain how neurons are activated in this processing. There are two major findings.

\begin{itemize}
\item LLMs can benefit from receiving parallel input in multiple languages. By transforming single-language input into multi-language input, we build a multi-source LLM that uses contexts from all these languages to make predictions. On the FLORES-200 machine translation benchmark, it achieves improvements of 11.3 BLEU points and 1.52 COMET points over the baseline.
\item Somewhat surprisingly, as more languages are involved in the input, fewer neurons are activated in the LLMs, facilitating more targeted and effective neuron activation patterns. This result links multilingual representation learning to \textit{synaptic pruning} in neuroscience \cite{huttenlocher1979synaptic,huttenlocher1990morphometric}: as a brain develops, some neural connections are strengthened, while others are deemed redundant and eliminated, making the transmission of neural signals more efficient.
\end{itemize}

More specifically, we find that in addition to the performance improvements from incorporating more languages, LLMs can gain advantages from extensive languages even involving ones that do not surpass baseline performances. With the help of high-quality machine translation, we efficiently acquire abundant parallel input, enabling us to apply this method to various tasks. Experimental results across 8 datasets, 7 languages, and 10 LLMs further demonstrate the effectiveness and applicability of PMI.

Since previous neuron activation statistics are primarily designed for the vanilla transformer model \cite{DBLP:conf/acl/ZhangL00S022,DBLP:conf/iclr/LiYBLRRYCYGK23}, we have extended these methods to analyze more advanced LLM architectures. When LLMs receive PMI, we observe simultaneous performance improvements and neuron inhibition. In addition, PMI selectively activates only a small portion of the most commonly used neurons while inhibiting the rest. Further analysis reveals that few-shot learning produces a similar effect on neuron activation, and integrating it with PMI enhances this neural reaction. These findings are consistently sustained across different models and tasks.

We introduce our PMI and evaluate it with human translation in Section \ref{sec:s1}. Subsequently, we comprehensively analyze the performance gains brought by PMI in Section \ref{sec:s2} and explain its effectiveness from a view of neuron activation in Section \ref{sec:s3}. Moreover, we apply PMI to various tasks under real scenario setups in Section \ref{sec:s4}.

\begin{figure}
\centering
\includegraphics[width=0.48\textwidth]{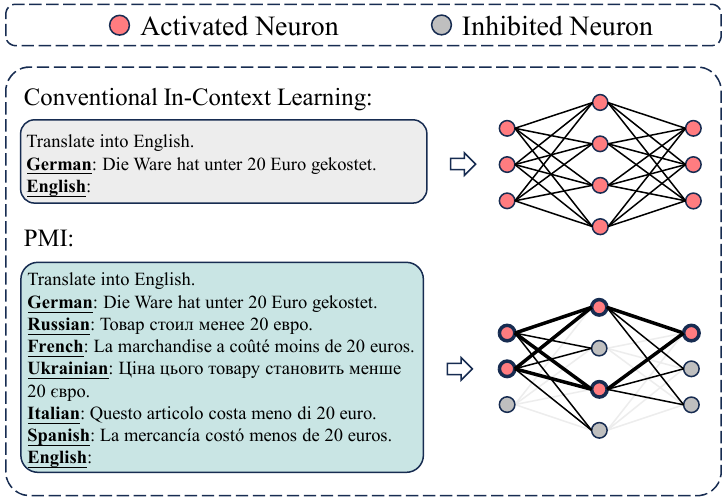}
\vspace{-1.5em}
\caption{Compared to conventional ICL, PMI inhibits neurons and promotes more precise activation (i.e., the thickened line). Other prompts are shown in Table \ref{app tab: All the prompts used in experiments}.}
\label{fig:f2}
\end{figure}

\section{Parallel Multilingual Input}

\subsection{LLMs benefit from PMI}
\label{sec:s1}

Given an input $\mathbf{X}$ of a task and a template $f(\cdot)$ to transform the input to an instruction, the conventional ICL can be expressed as follows:
\begin{eqnarray}
\mathbf{Y} & = & \mathrm{argmax} ~ P(y_{t}|f(\mathbf{X}))
\end{eqnarray}
where $\mathbf{Y}$ denotes the target output of the task and $y_t$ denotes the token generated at moment $t$. PMI extends beyond the conventional ICL approach of feeding LLMs solely with inputs in one language. Instead, it encompasses providing input in multiple languages, translated by professional human translators or sophisticated machine translation (MT) systems. The PMI can be shown as:
\begin{eqnarray}
\mathbf{Y} & = & \mathrm{argmax} ~ P(y_{t}|f(\mathbf{M},\mathbf{X}))
\end{eqnarray}
where $\mathbf{M}=\{m_1, m_2, ..., m_k\}$ is a parallel language set containing $k$ translations of the input. The template $f(\cdot)$ we used is neutral for both the input $\mathbf{X}$ and its translations $\mathbf{M}$, making LLMs cannot distinguish them. Figure \ref{fig:f2} shows the difference between the conventional ICL and our PMI when translating De $\rightarrow$ En.

Three aspects should be considered when constructing a PMI prompt: the choice of languages, the choice of translators, and the display order of languages. As shown in Appendix \ref{app:sec1}, our preliminary experiments suggest that: (1) choosing the language that LLMs understand better is crucial; (2) higher translation quality can lead to larger improvements; (3) it is preferable to place languages better understood at head and tail of the input sequence.

\begin{table}[t!]
\LARGE
    \centering

\resizebox{1.0\linewidth}{!}{
    \begin{tabular}{llcccc}
        \toprule

        \multirow{2.5}{*}{\textbf{Method}} & \multirow{2.5}{*}{\textbf{Input}} & \multicolumn{2}{c}{\textbf{ChatGPT}} & \multicolumn{2}{c}{\textbf{Qwen-14B}} \\
        \cmidrule(lr){3-4} \cmidrule(lr){5-6}
         & & \textbf{BLEU} & \textbf{COMET} & \textbf{BLEU} & \textbf{COMET}\\       

        \midrule
        \multicolumn{6}{c}{\textit{German $\rightarrow$ English}} \\
        \midrule
        \textbf{Direct} & De & 44.3 & 89.8 & 45.2 & 89.5 \\
        \multirow{2}{*}{\textbf{Pivot}} & Fr & 45.6 & 89.6 & 47.2 & 89.6 \\
         & Ru & 35.2 & 87.0 & 37.1 & 86.9 \\
        \rowcolor{shadow} \textbf{\textsc{PMI}-1} & De + Ru & 46.2 & 90.0 & 47.9 & 90.0 \\
        \rowcolor{shadow} \textbf{\textsc{PMI}-3} & De + Ru + Fr + Uk & 49.2 & 90.4 & 56.2 & 90.9 \\
        \rowcolor{shadow} \textbf{\textsc{PMI}-5} & De + Ru + Fr + Uk + It + Es & \textbf{50.2} & \textbf{90.6} & \textbf{56.5} & \textbf{91.0} \\

        \midrule
        \multicolumn{6}{c}{\textit{English $\rightarrow$ German}} \\
        \midrule
        \textbf{Direct} & En & \textbf{40.5} & 88.8 & \textbf{35.0} & 87.2 \\
        \multirow{2}{*}{\textbf{Pivot}} & Fr & 30.4 & 86.5 & 25.9 & 84.7	\\
         & Ru & 25.8 & 85.2 & 22.6 & 83.4 \\
        \rowcolor{shadow} \textbf{\textsc{PMI}-1} & En + Ru & 40.1 & 88.8 & 34.4 & 87.2	 \\
        \rowcolor{shadow} \textbf{\textsc{PMI}-3} & En + Ru + Fr + Uk & 40.3 & 88.8 & 34.8 & 87.4	 \\
        \rowcolor{shadow} \textbf{\textsc{PMI}-5} & En + Ru + Fr + Uk + It + Es & \textbf{40.5} & \textbf{88.9} & 34.6 & \textbf{87.5} \\

        \midrule
        \multicolumn{6}{c}{\textit{German $\rightarrow$ French}} \\
        \midrule
        \textbf{Direct} & De & 37.2 & 86.2 & 35.2 & 85.3 \\
        \multirow{2}{*}{\textbf{Pivot}} & Ro & 39.6 & 87.4 & 37.2 & 86.2 \\
         & Ru & 29.5 & 84.0 & 30.7 & 83.6 \\
        \rowcolor{shadow} \textbf{\textsc{PMI}-1} & De + Ru & 39.3 & 86.7 & 36.6 & 85.7	\\
        \rowcolor{shadow} \textbf{\textsc{PMI}-3} & De + Ru + Ro + Uk & 41.4 & 87.1 & 40.7 & 86.5 \\
        \rowcolor{shadow} \textbf{\textsc{PMI}-5} & De + Ru + Ro + Uk + It + Es & \textbf{42.4} & \textbf{87.3} & \textbf{42.3} & \textbf{86.9} \\    

        \bottomrule
    \end{tabular} 
    }

    \caption{Experiments of \textsc{PMI}, direct and pivot translation on the FLORES-200. We provide $k$ parallel languages denoted as \textsc{PMI}-$k$. Pivot row reports the best performance among all pivot translations in the first line and the performance of Russian in the second line.}

    \label{tab:t1}
    

\end{table}

\paragraph{Experimental Settings.}
We conducted translation experiments on the FLORES-200, which allowed us to probe the upper bound of the performance by constructing PMI using human-translated parallel sentences. Direct and pivot translation were our baselines. We utilized two powerful multilingual LLMs, including ChatGPT (\texttt{gpt-3.5-turbo-0613}) and Qwen-14B (\texttt{Qwen-14B-Chat}) \cite{DBLP:journals/corr/abs-2309-16609} \footnote{We also tried Bloomz \cite{DBLP:conf/acl/MuennighoffWSRB23}, however, compared to the performance on WMT, it showed deviantly high performance on FLORES-200, indicating a data leakage, which is also reported by \citet{DBLP:journals/corr/abs-2304-04675}.}. ChatGPT was prompted with one-shot for baseline and PMI prompts. While Qwen-14B exhibited confusion when processing PMI prompts, so we made some instruction training data of PMI and baseline prompts, and employed the LoRA technique \cite{DBLP:conf/iclr/HuSWALWWC22} to fine-tune Qwen-14B. More details can be found in Appendix \ref{app sec: Details of Experiment Setups}. The translation performance was evaluated in terms of SacreBLEU \cite{DBLP:conf/wmt/Post18} and COMET-22 (\texttt{wmt22-comet-da}) \cite{DBLP:conf/wmt/ReiSAZFGLCM22}.

\paragraph{Results and Analyses.}
Table \ref{tab:t1} delineates the performance of direct translation (Direct), pivot translation (Pivot), and PMI in three translation directions. We see, first of all, PMI achieves the best result among all the baselines, especially when more parallel languages are used. Despite the fact that the COMET score of some baselines reaches as high as 90, PMI still beats both direct and pivot translation with significant improvements. Furthermore, we find that PMI even benefits from parallel languages, which perform worse than direct translation. For example, integrating Russian into PMI achieves better performance than the baseline. Besides, when English becomes the original input, PMI leads to a small performance increase. We attribute this to the fact that LLMs have shown great success in understanding English input, leaving little room for improvement.

\subsection{Multiple Languages or Information Sources?}
\label{sec:s2}

\begin{figure*}
\centering
\includegraphics[width=0.90\textwidth]{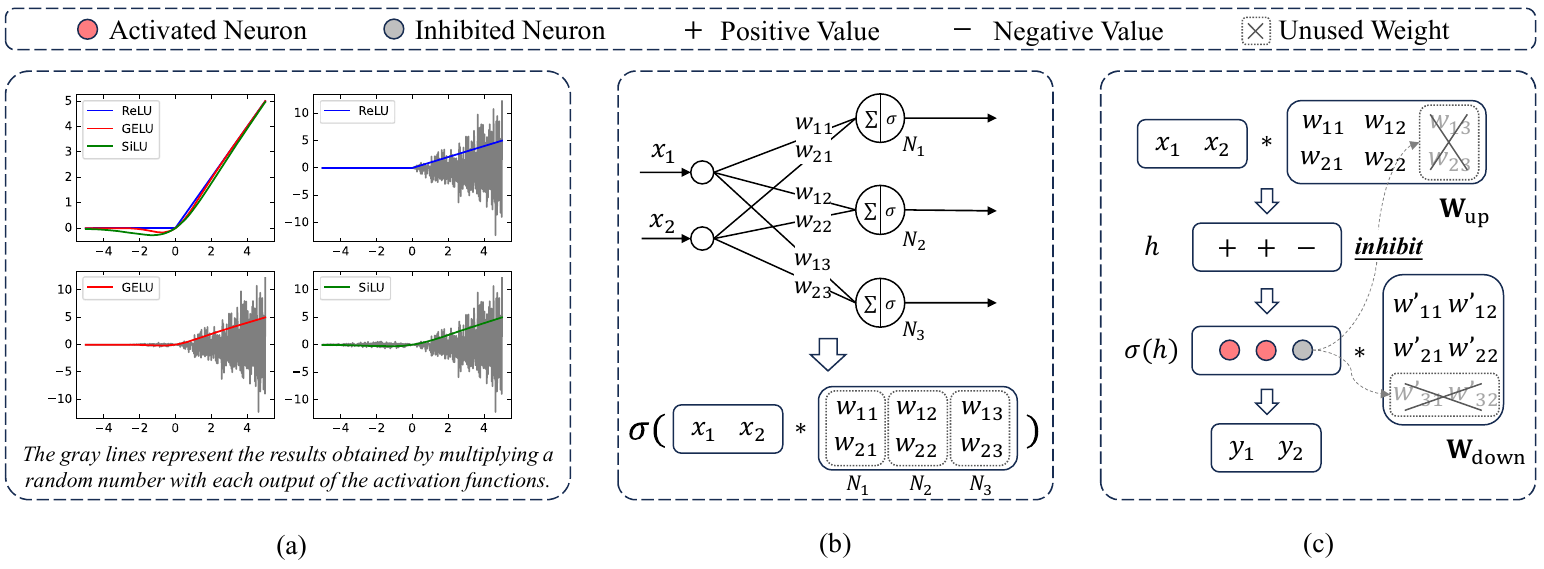}
\vspace{-0.4em}
\caption{The impact of ReLU-like activation functions on neurons during the forward process of transformer models. Figure (a) shows that activation function $\sigma(\cdot)$ like ReLU and some of its variants, when encountering negative inputs, saturate to zero and weaken the values multiplied by their outputs. Figure (b) details the equivalence between artificial neurons and the linear-transform matrix of MLPs. Figure (c) illustrates that ReLU-like activation functions inhibit neurons in $\mathbf{W}_{\mathrm{up}}$ and some weights of $\mathbf{W}_{\mathrm{down}}$ when the input is negative.}
\label{fig:f3}
\end{figure*}

Due to the parallel languages being translated by numerous human experts in the above experiments, one may argue that the improvement of PMI results from multiple information sources rather than languages. Specifically, multiple information sources can bring different perspectives of the original input, and translating inputs derived from human experts is like doing ensemble learning based on various strong translation systems. To separately quantify the effects of multiple languages and information sources, we decompose the PMI based on the human translations ($\textsc{PMI}_{GT}$) into three prompting strategies:
\begin{itemize}
    \item \textbf{Mono-source and monolingual:} The original input is paraphrased into different versions without changing the semantics. We denote this prompt as $\textsc{PMI}_{PA}$.
    \item \textbf{Multi-source but monolingual:} The human translation texts used in PMI are translated into the language of the original input by one translator. This prompt integrates different information sources but expresses them in one language, e.g., we provide ``De + De (Ru) + De (Fr) + De (Uk) + De (It) + De (Es)'' to LLMs where the language in parentheses represents the human translation text. We call it $\textsc{PMI}_{MS}$.
    \item \textbf{Multilingual but mono-source:} The original input is translated into different parallel languages by one translator. The source of this prompt is only the original input whereas the expression holds a multilingual form, like ``De + Ru (De) + Fr (De) + Uk (De) + It (De) + Es (De)'', which is represented by $\textsc{PMI}_{ML}$. We also illustrate these prompts in Figure \ref{fig:illustration of ablation study prompts}.
\end{itemize}

\begin{table}[t!]
\Large
    \centering

 \resizebox{1.0\linewidth}{!}{
    \begin{tabular}{ll|ll|ll}
        \toprule
        \multicolumn{2}{c|}{\textbf{System}} & \textbf{BLEU} & \textbf{COMET} & \textbf{BLEU} & \textbf{COMET} \\
        \midrule
        \multicolumn{2}{c|}{\textbf{\textit{Direction}}} & \multicolumn{2}{c|}{\textit{De $\rightarrow$ En}}  & \multicolumn{2}{c}{\textit{De $\rightarrow$ Fr}}  \\

        \midrule
        \multirow{5}{*}{\textbf{ChatGPT}} & Direct & \textbf{44.3}  & \textbf{89.8} & 37.2  & 86.2  \\
         & $\textsc{PMI}_{PA}$ & 36.4$^{\downarrow 7.9}$  & 88.6$^{\downarrow 1.1}$ & 34.8$^{\downarrow 2.4}$  & 85.5$^{\downarrow 0.7}$  \\
         & $\textsc{PMI}_{MS}$ & 42.6$^{\downarrow 1.7}$  & 89.4$^{\downarrow 0.3}$ & 37.1$^{\downarrow 0.1}$  & 86.0$^{\downarrow 0.2}$ \\
         & $\textsc{PMI}_{ML}$ & 44.1$^{\downarrow 0.2}$  & 89.7$^{\downarrow 0.1}$ & \textbf{39.7}$^{\uparrow 2.5}$  & \textbf{86.6}$^{\uparrow 0.4}$ \\
         & $\textsc{PMI}_{GT}$ & \textit{50.2}  & \textit{90.6} & \textit{42.4}  & \textit{87.3} \\

        \midrule
        \multirow{5}{*}{\textbf{Qwen-14b}} & Direct & 45.5  & 89.6 & 35.4  & 85.4  \\
         & $\textsc{PMI}_{PA}$ & 40.4$^{\downarrow 5.1}$  & 89.0$^{\downarrow 0.6}$ & 31.8$^{\downarrow 3.6}$  & 84.6$^{\downarrow 0.8}$  \\
         & $\textsc{PMI}_{MS}$ & \textbf{46.6}$^{\uparrow 1.1}$  & \textbf{90.0}$^{\uparrow 0.4}$ & 36.5$^{\uparrow 1.1}$  & \textbf{86.1}$^{\uparrow 0.7}$  \\
         & $\textsc{PMI}_{ML}$ & 44.9$^{\downarrow 0.6}$  & 89.6$^{\uparrow 0.0}$ & \textbf{37.6}$^{\uparrow 2.2}$  & 86.0$^{\uparrow 0.6}$ \\
         & $\textsc{PMI}_{GT}$ & \textit{56.3}  & \textit{91.1} & \textit{42.8}  & \textit{87.0} \\

        \midrule
        \multirow{4}{*}{\textbf{GPT-4}} & Direct & 44.9  & \textbf{89.9} & 39.0  & 86.5\\
         & $\textsc{PMI}_{MS}$ & 43.6$^{\downarrow 1.3}$  & 89.8$^{\downarrow 0.1}$ & 39.6$^{\uparrow 0.6}$  & \textbf{87.0}$^{\uparrow 0.5}$ \\
         & $\textsc{PMI}_{ML}$ & \textbf{45.4}$^{\uparrow 0.5}$  & 89.7$^{\downarrow 0.1}$ & \textbf{40.1}$^{\uparrow 1.1}$  & 86.8$^{\uparrow 0.2}$  \\
         & $\textsc{PMI}_{GT}$ & \textit{52.9}  & \textit{90.9} & \textit{45.9}  & \textit{88.1} \\

        \bottomrule
    \end{tabular} 
    }

    \caption{The ablation study of the mono-source and monolingual ($\textsc{PMI}_{PA}$), multi-source but monolingual ($\textsc{PMI}_{MS}$), multilingual but mono-source ($\textsc{PMI}_{ML}$), multi-source and multilingual ($\textsc{PMI}_{GT}$) prompts on the FLORES-200. The best results are in bold among all the prompts except for $\textsc{PMI}_{GT}$.}

    \label{tab:t2}
    

\end{table}

\paragraph{Experimental Settings.}
With access to Qwen-14B, ChatGPT, and GPT-4 (\texttt{gpt-4-0613}), we conducted experiments on two translation directions of FLORES-200. The translation system used by both $\textsc{PMI}_{MS}$ and $\textsc{PMI}_{ML}$ prompt was the NLLB-54B model \cite{DBLP:journals/corr/abs-2207-04672}. We derived the paraphrased sentences by requesting ChatGPT. Notably, Qwen-14B used in this experiment is different from the one in the previous experiment, as we have to fine-tune Qwen-14B with extra training data based on the $\textsc{PMI}_{MS}$ prompt for fairness.

\paragraph{Results and Analyses.}
From Table \ref{tab:t2}, we can see that both $\textsc{PMI}_{MS}$ and $\textsc{PMI}_{ML}$ prompt achieve improvement most of the time, while none of them can reach the same performance as the $\textsc{PMI}_{GT}$ prompt. In addition, the $\textsc{PMI}_{ML}$ prompt far outperforms the $\textsc{PMI}_{PA}$ prompt, which demonstrates that multilingual input helps LLMs again. Also, we see that despite the similar baseline performance, GPT-4 always outperforms ChatGPT significantly when being armed with PMI, suggesting that stronger LLMs benefit more from the PMI.

\begin{figure*}
\centering
\begin{tikzpicture}[scale=0.42]
\tikzstyle{every node}=[font=\Large]

    \node [rectangle,draw=black,minimum height=1.0em,minimum width=36.43em,font=\small,anchor=west,align=center] (final) at (0em,18em){};
    
    \node [anchor=north,align=center,minimum size=1pt,scale=0.50] (a) at (2.7,7.4){ \ref{pgfplots:triangle}};
    \node [anchor=center] at (4.9,6.9){\scriptsize Activation};

    \node [anchor=north,align=center,minimum size=1pt,scale=0.50] (a) at (9.2,7.4){ \ref{pgfplots:circle}};
    \node [anchor=center] at (11.4,6.9){\scriptsize Inhibition};  
    
    \node [rectangle,draw=none,fill=cGray,inner sep=2pt,minimum height=0.6em,minimum width=2em,font=\small,anchor=north,align=center] (a) at (16.3,7.2){};
    \node [anchor=center] at (20.0,6.9){\scriptsize Single Language};
    
    \node [rectangle,draw=none,fill=cGreen,inner sep=2pt,minimum height=0.6em,minimum width=2em,font=\small,anchor=north,align=center] (a) at (25.5,7.2){};
    \node [anchor=center] at (29.2,6.9){\scriptsize Multiple Languages};

    \begin{axis}[
      x=4em,
      at={(0,0)},
      ybar,
      bar width=.58cm,
      bar shift=0pt,
      axis y line*=left,
      ymin=88,
      xtick={1,2,3,4},
      x tick style={draw=none},
      xticklabels={Direct,\textsc{PMI}-1,\textsc{PMI}-3,\textsc{PMI}-5},
      xticklabel style={rotate=20, anchor=center, xshift=-0.5em, yshift=-1.5em},
      xlabel=(a) $De \rightarrow En$$^\ast$,
      x label style={yshift=-1.0em},
      ylabel=\ref{pgfplots:gray bar} \ref{pgfplots:green bar}  COMET,
      y tick label style={
        /pgf/number format/.cd,
            fixed,
            fixed zerofill,
            precision=1,
        /tikz/.cd
        },
    ]
        \addplot [
            draw = none,
            fill = cGray
        ]  coordinates {(1,89.5)};
        \label{pgfplots:gray bar}
        \addplot [
            draw = none,
            fill = cGreen
        ]  coordinates {(2,90.0) (3,90.9) (4,91.1)};
        \label{pgfplots:green bar}
    \end{axis}
    
    \begin{axis}[
      x=4em,
      at={(0,0)},
      axis y line*=right,
      axis x line=none,
      xtick=data,
      xticklabel style={rotate=20, anchor=center, xshift=0em, yshift=-1.0em},
      symbolic x coords={Direct,\textsc{PMI}-1, \textsc{PMI}-3, \textsc{PMI}-5},
      y tick label style={
        /pgf/number format/.cd,
            fixed,
            fixed zerofill,
            precision=1,
        /tikz/.cd
        },
    ]
        \addplot [
            ultra thick, color = cRed, mark=*,
        ]  coordinates {(Direct,28.054)(\textsc{PMI}-1,27.983) (\textsc{PMI}-3,27.863) (\textsc{PMI}-5,27.789)};
        \addplot [
            only marks, ultra thick, color = cRed, mark=o, mark size=8pt,
        ]  coordinates {(\textsc{PMI}-1,27.983) (\textsc{PMI}-3,27.863) (\textsc{PMI}-5,27.789) };
        \label{pgfplots:circle}
    \end{axis}
    \draw [thick, cRed, dashed] (0,5.2)--(5.57,5.2);

    \begin{axis}[
      x=4em,
      at={(22em,0)},
      ybar,
      bar width=.58cm,
      bar shift=0pt,
      axis y line*=left,
      ymin=84,
      xtick={1,2,3,4,5},
      x tick style={draw=none},
      xticklabels={Direct,\textsc{PMI}-1,\textsc{PMI}-3,\textsc{PMI}-5},
      xticklabel style={rotate=20, anchor=center, xshift=-0.5em, yshift=-1.5em},
      xlabel=(b) $De \rightarrow Fr$$^\ast$,
      x label style={yshift=-1.0em},
      y tick label style={
        /pgf/number format/.cd,
            fixed,
            fixed zerofill,
            precision=1,
        /tikz/.cd
        },
    ]
        \addplot [
            draw = none,
            fill = cGray
        ]  coordinates {(1,85.3)};
        \addplot [
            draw = none,
            fill = cGreen
        ]  coordinates {(2,85.7) (3,86.5) (4,86.9)};
    \end{axis}
    
    \begin{axis}[
      x=4em,
      at={(22em,0)},
      axis y line*=right,
      axis x line=none,
      xtick=data,
      xticklabel style={rotate=20, anchor=center, xshift=0em, yshift=-1.0em},
      xlabel=$x\_label$,
      ytick={27.1, 27.2},
      symbolic x coords={Direct, \textsc{PMI}-1, \textsc{PMI}-3, \textsc{PMI}-5},
      y tick label style={
        /pgf/number format/.cd,
            fixed,
            fixed zerofill,
            precision=1,
        /tikz/.cd
        },
    ]
        \addplot [
            ultra thick, color = cRed, mark=*,
        ]  coordinates {(Direct,27.226) (\textsc{PMI}-1,27.186) (\textsc{PMI}-3,27.152) (\textsc{PMI}-5,27.070)};
        \addplot [
            only marks, ultra thick, color = cRed, mark=o, mark size=8pt,
        ]  coordinates {(\textsc{PMI}-1,27.186) (\textsc{PMI}-3,27.152) (\textsc{PMI}-5,27.070)};
    \end{axis}
    \draw [thick, cRed, dashed] (22em,5.2)--(36.5em,5.2);

    \begin{axis}[
      at={(44em,0)},
      ybar,
      bar width=.58cm,
      bar shift=0pt,
      axis y line*=left,
      ymin=88,
      xtick={1,2,3,4,5},
      x tick style={draw=none},
      xticklabels={Direct,$\textsc{PMI}_{PA}$,$\textsc{PMI}_{MS}$,$\textsc{PMI}_{ML}$,$\textsc{PMI}_{GT}$},
      xticklabel style={rotate=20, anchor=center, xshift=-0.5em, yshift=-1.5em},
      xlabel=(c) $De \rightarrow En$$^\dag$,
      x label style={yshift=-0.5em},
      y tick label style={
        /pgf/number format/.cd,
            fixed,
            fixed zerofill,
            precision=1,
        /tikz/.cd
        },
    ]
        \addplot [
            draw = none,
            fill = cGray
        ]  coordinates {(1,89.6) (2,89.0) (3,90.0)};
        \addplot [
            draw = none,
            fill = cGreen
        ]  coordinates {(4,89.6) (5,91.1)};
    \end{axis}
    
    \begin{axis}[
      at={(44em,0)},
      axis y line*=right,
      axis x line=none,
      xtick=data,
      xticklabel style={rotate=20, anchor=center, xshift=0em, yshift=-1.0em},
      xlabel=$x\_label$,
      symbolic x coords={Direct,$\textsc{PMI}_{PA}$,$\textsc{PMI}_{MS}$,$\textsc{PMI}_{ML}$,$\textsc{PMI}_{GT}$},
      y tick label style={
        /pgf/number format/.cd,
            fixed,
            fixed zerofill,
            precision=1,
        /tikz/.cd
        },
    ]
        \addplot [
            ultra thick, color = cRed, mark=*,
        ]  coordinates {(Direct,28.068) ($\textsc{PMI}_{PA}$,28.109)($\textsc{PMI}_{MS}$,28.148) ($\textsc{PMI}_{ML}$,27.958)  ($\textsc{PMI}_{GT}$,27.851)};
        \addplot [
            only marks, ultra thick, color = cRed, mark=triangle, mark size=10pt,
        ]  coordinates {($\textsc{PMI}_{PA}$,28.109) ($\textsc{PMI}_{MS}$,28.148)};
        \label{pgfplots:triangle}
        \addplot [
            only marks, ultra thick, color = cRed, mark=o, mark size=8pt,
        ]  coordinates {($\textsc{PMI}_{ML}$,27.958) ($\textsc{PMI}_{GT}$,27.851)};
    \end{axis}
    \draw [thick, cRed, dashed] (44em,4.0)--(62.2em,4.0);

    \begin{axis}[
      at={(69em,0)},
      ybar,
      bar width=.58cm,
      bar shift=0pt,
      axis y line*=left,
      ymin=83.5,
      xtick={1,2,3,4,5},
      x tick style={draw=none},
      xticklabels={Direct,$\textsc{PMI}_{PA}$,$\textsc{PMI}_{MS}$,$\textsc{PMI}_{ML}$,$\textsc{PMI}_{GT}$},
      xticklabel style={rotate=20, anchor=center, xshift=-0.5em, yshift=-1.5em},
      xlabel=(d) $De \rightarrow Fr$$^\dag$,
      x label style={yshift=-0.5em},
      y tick label style={
        /pgf/number format/.cd,
            fixed,
            fixed zerofill,
            precision=1,
        /tikz/.cd
        },
    ]
        \addplot [
            draw = none,
            fill = cGray
        ]  coordinates {(1,85.4) (2,84.6) (3,86.1)};
        \addplot [
            draw = none,
            fill = cGreen
        ]  coordinates {(4,86.0) (5,87.0)};
    \end{axis}
    
    \begin{axis}[
      at={(69em,0)},
      axis y line*=right,
      axis x line=none,
      xtick=data,
      xticklabel style={rotate=20, anchor=center, xshift=0em, yshift=-1.0em},
      xlabel=$x\_label$,
      ylabel=\ref{pgfplots:red line} Activation Proportion (\%),
      symbolic x coords={Direct,$\textsc{PMI}_{PA}$,$\textsc{PMI}_{MS}$,$\textsc{PMI}_{ML}$,$\textsc{PMI}_{GT}$},
      y tick label style={
        /pgf/number format/.cd,
            fixed,
            fixed zerofill,
            precision=1,
        /tikz/.cd
        },
    ]
        \addplot [
            ultra thick, color = cRed, mark=*,
        ]  coordinates {(Direct,27.196) ($\textsc{PMI}_{PA}$,27.380) ($\textsc{PMI}_{MS}$,27.381) ($\textsc{PMI}_{ML}$,27.165) ($\textsc{PMI}_{GT}$,27.108)};
        \label{pgfplots:red line}
        \addplot [
            only marks, ultra thick, color = cRed, mark=triangle, mark size=10pt,
        ]  coordinates {($\textsc{PMI}_{PA}$,27.380) ($\textsc{PMI}_{MS}$,27.381)};
        \addplot [
            only marks, ultra thick, color = cRed, mark=o, mark size=8pt,
        ]  coordinates {($\textsc{PMI}_{ML}$,27.165) ($\textsc{PMI}_{GT}$,27.108)};
    \end{axis}
    \draw [thick, cRed, dashed] (69em,2.0)--(87.2em,2.0);

\end{tikzpicture}
\vspace{-0.5em}
\caption{The COMET score and the activation proportion of Qwen-14B armed with different prompts on FLORES-200. Notably, whether a method inhibits or activates neurons depends on its activation proportion being below or above the baseline level. Thus, a point on the curves suggests inhibition \begin{small}{\textcolor{cRed}{$\bigcirc$}}\end{small} if it falls below the first point, and activation \textcolor{cRed}{$\bigtriangleup$} if it exceeds the first point. $\ast$ and $\dag$ indicates the model used in Section \ref{sec:s1} and \ref{sec:s2}, respectively.}
\label{fig:f4}
\end{figure*}
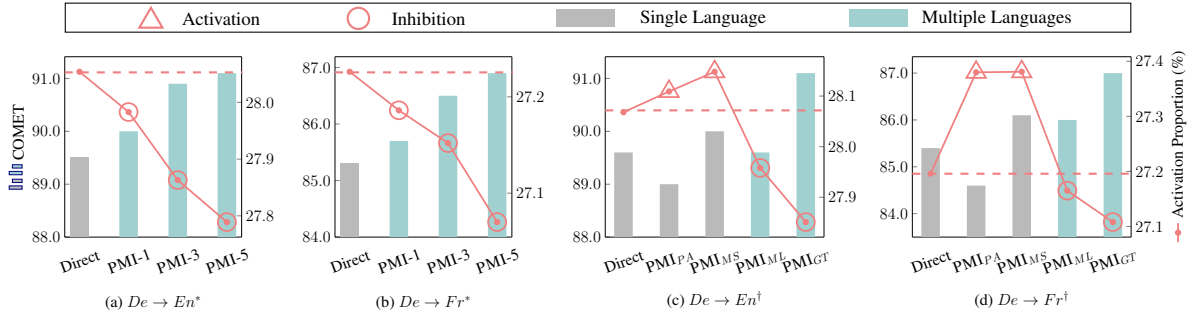

\begin{figure}
\centering
\includegraphics[width=0.47\textwidth]{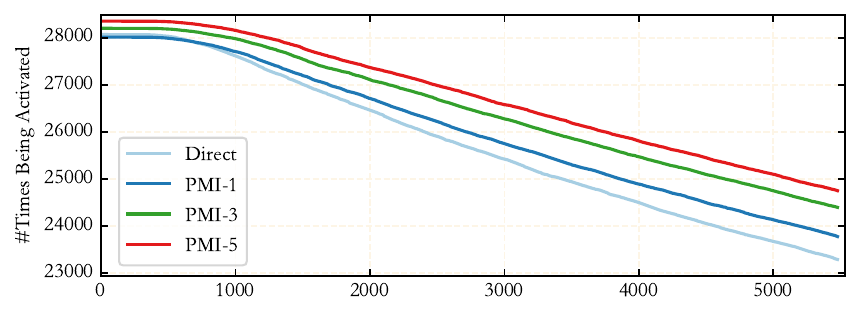}
\vspace{-0.5em}
\caption{The distribution of the top 1\% of activated neurons in Qwen-14B on FLORES-200 De $\rightarrow$ En. The horizontal axis represents different neurons arranged in descending order based on the number of times they are activated.}
\label{fig:f5}
\end{figure}

\section{PMI Can Help: From a View of Neuron Activation}
\label{sec:s3}
Although LLMs benefit from PMI, there is still no idea about how this mechanism works. Considering that knowledge is memorized in different neurons in transformer models \cite{DBLP:conf/acl/DaiDHSCW22}, a straightforward hypothesis is that giving the input in multiple languages may increase the number of activated neurons in the inference process. To quantify how many neurons in transformer models are activated during inference, some works propose to make statistics of the nonzero values in the intermediate output of multi-layer perceptrons (MLPs) after a ReLU activation function \cite{DBLP:conf/acl/ZhangL00S022,DBLP:conf/iclr/LiYBLRRYCYGK23}. This is based on the idea that, in matrix multiplication, zero can be omitted; therefore, neurons that output zero are considered inhibited while others are activated. Next, we will explain this statistical method.

\subsection{Method of Counting Activated Neurons}

\paragraph{ReLU controls the life and death of neurons.}
In transformer models, the activation function $\sigma(\cdot)$ lies in the middle of the two-layer MLPs, like this:
\begin{eqnarray}
\mathbf{Y} = \sigma\left(\mathbf{X} \mathbf{W}_{\mathrm{up}}\right) \mathbf{W}_{\mathrm{down}}
\end{eqnarray}
where $\mathbf{X}$ and $\mathbf{Y}$ stand for input and output, respectively. $\mathbf{W}_{\mathrm{up}}$ and $\mathbf{W}_{\mathrm{down}}$ represent different MLP layers containing artificial neurons. The vanilla transformer uses ReLU as the activation function \cite{DBLP:conf/nips/VaswaniSPUJGKP17}, i.e., $\mathrm{max}(x,0)$. In Figure \ref{fig:f3} (b) and (c), ReLU outputs zero value means two aspects: the neuron in $\mathbf{W}_{\mathrm{up}}$ is inhibited and stripped from the whole neural network; the weight in $\mathbf{W}_{\mathrm{down}}$ that accepts the zero value is inhibited.

\paragraph{Counting activated neurons in MLPs with ReLU variants.}
Despite the success of ReLU, recent works find that making a ReLU-like non-linearity to output negative values can increase training speed \cite{DBLP:journals/corr/ClevertUH15,hendrycks2016gaussian}. Hence, as shown in Table \ref{app tab: activation functions}, these variants of ReLU become popular among LLMs. We draw ReLU, GELU and SiLU in Figure \ref{fig:f3} (a). We see that despite both GELU and SiLU performing as smooth ReLU, they retain their basic character, i.e., saturating to zero at negative input values and protecting positive input values. In other words, these ReLU variants significantly reduce the absolute value of any negative input to a level that is close to or equal to zero. As a result, some neurons and weights are inhibited as before. This motivates us to make statistics of activated neurons in MLPs with ReLU variants by \textit{counting the output values of the activation function that are greater than zero}.

Other works combine GELU and SiLU with the gated linear units \cite{DBLP:journals/corr/abs-2002-05202} like this:
\begin{eqnarray}
\mathbf{Y} = \left(\sigma\left(\mathbf{X} \mathbf{W}_{\mathrm{up}}\right) \odot \left(\mathbf{X} \mathbf{V}_{\mathrm{up}}\right)\right) \mathbf{W}_{\mathrm{down}}
\end{eqnarray} 
where $\odot$ is the element-wise product and a new matrix $V_{\mathrm{up}}$ is introduced to perform the gate. If we transform the formula into this:
\begin{eqnarray}
\mathbf{Y} = \sigma\left(\mathbf{X} \mathbf{W}_{\mathrm{up}}\right) {\left(\mathbf{X} \mathbf{V}_{\mathrm{up}} \odot {\mathbf{W}_{\mathrm{down}}}^\top \right)}^\top
\end{eqnarray}
then we can consider $\mathbf{X} \mathbf{V}_{\mathrm{up}} \odot {\mathbf{W}_{\mathrm{down}}}^\top$ as a whole, and both inhibiting neurons and weights happen as before. Thus, our statistical method of activated neurons remains unchanged.

\subsection{Experiments and Results}
Figure \ref{fig:f4} shows performances and the proportion of activated neurons\footnote{Note that the proportion mentioned is derived by averaging the percentages of activated neurons for each token generated by an LLM across the dataset. We discuss this implementation in detail in Appendix \ref{app sec: More explanations of counting activated neurons}.} on Qwen-14B models. From the results, we get the following observations:

\paragraph{Activated neurons are far fewer than inhibited ones.} Despite performing dense computations, only a small number of neurons, around $27\%$, are activated in Qwen-14B during the inference stage, which is similar to the sparse activation phenomenon observed by \citet{DBLP:conf/iclr/LiYBLRRYCYGK23}. Besides, the differences in the proportion of activated neurons are small in numerical terms, we attribute this to the finding that few parameters are in charge of linguistic competence in LLMs \cite{DBLP:journals/corr/abs-2310-14928}.

\paragraph{More languages, more inhibited neurons, more performance gain.} As shown in Figure \ref{fig:f4} (a) and (b), if we add more parallel languages in PMI, then the proportion of activated neurons becomes small meanwhile LLM yields better translations, indicating a consistent correlation between inhibiting neurons and performance improvements.

\begin{table*}[t!]
\Large
    \centering

 \resizebox{0.95\linewidth}{!}{
    \begin{tabular}{ll|cc|cc|cc|cc|cc|cc}
        \toprule
        \multicolumn{2}{c|}{\textbf{System}} & \large{\textbf{BLEU}} & \large{\textbf{COMET}} & \large{\textbf{BLEU}} & \large{\textbf{COMET}} & \large{\textbf{BLEU}} & \large{\textbf{COMET}} & \large{\textbf{BLEU}} & \large{\textbf{COMET}} & \large{\textbf{BLEU}} & \large{\textbf{COMET}} & \large{\textbf{BLEU}} & \large{\textbf{COMET}}  \\
        \midrule
        \multicolumn{2}{c|}{\textbf{\textit{Direction}}} & \multicolumn{2}{c|}{\textit{De $\rightarrow$ En}} & \multicolumn{2}{c|}{\textit{Zh $\rightarrow$ En}}  & \multicolumn{2}{c|}{\textit{De $\rightarrow$ Fr}}  & \multicolumn{2}{c|}{\textit{En $\rightarrow$ De}} & \multicolumn{2}{c|}{\textit{En $\rightarrow$ Zh}} & \multicolumn{2}{c}{\textit{Is $\rightarrow$ En}} \\
        \midrule
        \multicolumn{2}{c|}{\textbf{\textit{Parallel Languages}}} & \multicolumn{2}{c|}{\large{\textit{Es Ru Fr Zh Ja Cs}}} & \multicolumn{2}{c|}{\large{\textit{Es Ru Fr Ja Cs De}}}  & \multicolumn{2}{c|}{\large{\textit{En Ru Es Zh It Cs}}}  & \multicolumn{2}{c|}{\large{\textit{Es Ru Fr Zh Ja Cs}}} & \multicolumn{2}{c|}{\large{\textit{Es Ru Fr Ja Cs De}}} & \multicolumn{2}{c}{\large{\textit{Es Ru Fr It Cs De}}} \\
        
        \midrule
        \multirow{5}{*}{\textbf{ChatGPT$^\ast$}} & Direct & 29.8 & 82.7 & \textbf{24.7} & 81.9 & 38.6 & 84.1 & \textbf{34.5} & 87.2 & \textbf{43.8} & \textbf{87.2} & 35.6 & 84.5  \\
         & Pivot & 28.5 & 84.0 & 21.6 & 81.9 & 40.4 & 84.0 & 30.0 & 86.4 & 40.3 & 86.0 & 35.0 & 85.6  \\
         & \cellcolor{shadow}\textsc{PMI}-1 & \cellcolor{shadow}\textbf{32.4} & \cellcolor{shadow}85.3 & \cellcolor{shadow}24.6 & \cellcolor{shadow}\textbf{82.8} & \cellcolor{shadow}40.9 & \cellcolor{shadow}\textbf{84.5} & \cellcolor{shadow}34.0 & \cellcolor{shadow}87.3 & \cellcolor{shadow}41.8 & \cellcolor{shadow}86.5 & \cellcolor{shadow}38.0 & \cellcolor{shadow}86.4  \\
         & \cellcolor{shadow}\textsc{PMI}-3 & \cellcolor{shadow}32.1 & \cellcolor{shadow}85.4 & \cellcolor{shadow}23.4 & \cellcolor{shadow}82.6 & \cellcolor{shadow}41.1 & \cellcolor{shadow}\textbf{84.5} & \cellcolor{shadow}\textbf{34.5} & \cellcolor{shadow}87.5 & \cellcolor{shadow}41.7 & \cellcolor{shadow}86.9 & \cellcolor{shadow}38.2 & \cellcolor{shadow}86.6  \\
          & \cellcolor{shadow}\textsc{PMI}-6 & \cellcolor{shadow}31.6 & \cellcolor{shadow}\textbf{85.5} & \cellcolor{shadow}18.6 & \cellcolor{shadow}82.4 & \cellcolor{shadow}\textbf{41.3} & \cellcolor{shadow}\textbf{84.5} & \cellcolor{shadow}\textbf{34.5} & \cellcolor{shadow}\textbf{87.6} & \cellcolor{shadow}41.7 & \cellcolor{shadow}86.9 & \cellcolor{shadow}\textbf{38.5} & \cellcolor{shadow}\textbf{86.7}  \\
        

        \midrule
        \multirow{5}{*}{\textbf{LLaMA3-8B$^\ast$}} & Direct & \textbf{30.4} & 84.0 & 21.4 & 80.2 & 29.2 & 79.8 & 27.3 & 83.2 & \textbf{35.8} & 83.7 & 22.1 & 76.7  \\
         & Pivot & 27.4 & 83.4 & 21.3 & 81.4 & 31.7 & 80.8 & 22.8 & 81.8 & 29.3 & 81.7 & 31.0 & 84.6  \\
        & \cellcolor{shadow}\textsc{PMI}-1 & \cellcolor{shadow}30.3 & \cellcolor{shadow}85.0 & \cellcolor{shadow}23.2 & \cellcolor{shadow}82.1 & \cellcolor{shadow}33.4 & \cellcolor{shadow}81.5 & \cellcolor{shadow}26.1 & \cellcolor{shadow}83.4 & \cellcolor{shadow}32.5 & \cellcolor{shadow}82.8 & \cellcolor{shadow}34.7 & \cellcolor{shadow}85.2  \\
        & \cellcolor{shadow}\textsc{PMI}-3 & \cellcolor{shadow}30.1 & \cellcolor{shadow}\textbf{85.1} & \cellcolor{shadow}23.4 & \cellcolor{shadow}82.4 & \cellcolor{shadow}33.9 & \cellcolor{shadow}82.3 & \cellcolor{shadow}\textbf{27.4} & \cellcolor{shadow}84.6 & \cellcolor{shadow}35.1 & \cellcolor{shadow}83.5 & \cellcolor{shadow}\textbf{36.6} & \cellcolor{shadow}\textbf{86.0}  \\
        & \cellcolor{shadow}\textsc{PMI}-6 & \cellcolor{shadow}29.9 & \cellcolor{shadow}\textbf{85.1} & \cellcolor{shadow}\textbf{24.1} & \cellcolor{shadow}\textbf{82.7} & \cellcolor{shadow}\textbf{34.5} & \cellcolor{shadow}\textbf{82.5} & \cellcolor{shadow}27.3 & \cellcolor{shadow}\textbf{84.9} & \cellcolor{shadow}34.1 & \cellcolor{shadow}\textbf{84.1} & \cellcolor{shadow}36.0 & \cellcolor{shadow}85.8  \\

        \midrule
        \multirow{5}{*}{\textbf{Qwen-14B$^\dag$}} & Direct & 30.4 & 84.4 & 23.7 & 80.8 & 34.2 & 81.9 & 29.6 & 85.3 & \textbf{45.2} & \textbf{87.6} & 18.4 & 69.7  \\
         & Pivot & 28.2 & 84.0 & 22.4 & 81.8 & 37.4 & 82.7 & 26.9 & 84.7 & 41.2 & 86.3 & 34.1 & 85.4  \\
        & \cellcolor{shadow}\textsc{PMI}-1 & \cellcolor{shadow}31.3 & \cellcolor{shadow}84.8 & \cellcolor{shadow}\textbf{24.3} & \cellcolor{shadow}\textbf{82.0} & \cellcolor{shadow}38.0 & \cellcolor{shadow}83.1 & \cellcolor{shadow}29.7 & \cellcolor{shadow}85.4 & \cellcolor{shadow}45.1 & \cellcolor{shadow}\textbf{87.6} & \cellcolor{shadow}35.6 & \cellcolor{shadow}85.1  \\
        & \cellcolor{shadow}\textsc{PMI}-3 & \cellcolor{shadow}\textbf{31.6} & \cellcolor{shadow}\textbf{84.9} & \cellcolor{shadow}23.5 & \cellcolor{shadow}\textbf{82.0} & \cellcolor{shadow}37.7 & \cellcolor{shadow}\textbf{83.4} & \cellcolor{shadow}\textbf{30.0} & \cellcolor{shadow}\textbf{85.8} & \cellcolor{shadow}44.9 & \cellcolor{shadow}\textbf{87.6} & \cellcolor{shadow}37.2 & \cellcolor{shadow}85.6  \\
        & \cellcolor{shadow}\textsc{PMI}-6 & \cellcolor{shadow}31.0 & \cellcolor{shadow}\textbf{84.9} & \cellcolor{shadow}22.0 & \cellcolor{shadow}81.3 & \cellcolor{shadow}\textbf{38.4} & \cellcolor{shadow}\textbf{83.4} & \cellcolor{shadow}29.9 & \cellcolor{shadow}85.5 & \cellcolor{shadow}\textbf{45.2} & \cellcolor{shadow}\textbf{87.6} & \cellcolor{shadow}\textbf{37.9} & \cellcolor{shadow}\textbf{85.7}  \\

         \midrule
        \multirow{5}{*}{\textbf{ALMA-13B$^\dag$}} & Direct & 28.1 & 83.8 & 21.6 & 79.6 & 27.1 & 79.2 & 29.6 & 85.5 & 36.9 & 85.8 & 34.0 & 85.8  \\
         & Pivot & 26.0 & 83.3 & 21.7 & 81.2 & 29.9 & 80.3 & 26.4 & 84.8 & 32.3 & 84.6 & 32.7 & 85.2  \\
        & \cellcolor{shadow}\textsc{PMI}-1 & \cellcolor{shadow}29.9 & \cellcolor{shadow}84.6 & \cellcolor{shadow}\textbf{23.8} & \cellcolor{shadow}\textbf{81.8} & \cellcolor{shadow}31.1 & \cellcolor{shadow}80.8 & \cellcolor{shadow}29.7 & \cellcolor{shadow}85.3 & \cellcolor{shadow}36.9 & \cellcolor{shadow}85.9 & \cellcolor{shadow}37.0 & \cellcolor{shadow}86.3  \\
        & \cellcolor{shadow}\textsc{PMI}-3 & \cellcolor{shadow}\textbf{30.8} & \cellcolor{shadow}\textbf{85.0} & \cellcolor{shadow}22.9 & \cellcolor{shadow}\textbf{81.8} & \cellcolor{shadow}\textbf{33.3} & \cellcolor{shadow}\textbf{81.5} & \cellcolor{shadow}\textbf{29.9} & \cellcolor{shadow}\textbf{86.0} & \cellcolor{shadow}36.9 & \cellcolor{shadow}\textbf{86.0} & \cellcolor{shadow}\textbf{38.3} & \cellcolor{shadow}\textbf{86.5}  \\
        & \cellcolor{shadow}\textsc{PMI}-6 & \cellcolor{shadow}30.0 & \cellcolor{shadow}84.9 & \cellcolor{shadow}18.1 & \cellcolor{shadow}79.5 & \cellcolor{shadow}\textbf{33.3} & \cellcolor{shadow}\textbf{81.5} & \cellcolor{shadow}\textbf{29.9} & \cellcolor{shadow}85.9 & \cellcolor{shadow}\textbf{37.2} & \cellcolor{shadow}\textbf{86.0} & \cellcolor{shadow}38.2 & \cellcolor{shadow}86.3  \\

         \midrule
        \multirow{5}{*}{\textbf{mT0-13B$^\ast$}} & Direct & 25.1 & 82.2 & 13.7 & 76.2 & 27.9 & 78.5 & \textbf{17.6} & 77.3 & 26.0 & 83.1 & 29.9 & 83.9  \\
         & Pivot & 24.5 & 82.5 & 19.3 & \textbf{80.7} & 30.5 & 80.0 & 17.4 & \textbf{78.5} & 23.8 & 82.1 & 30.8 & 84.6  \\
        & \cellcolor{shadow}\textsc{PMI}-1 & \cellcolor{shadow}27.0 & \cellcolor{shadow}83.4 & \cellcolor{shadow}18.3 & \cellcolor{shadow}79.9 & \cellcolor{shadow}29.9 & \cellcolor{shadow}79.4 & \cellcolor{shadow}17.4 & \cellcolor{shadow}76.5 & \cellcolor{shadow}25.5 & \cellcolor{shadow}82.4 & \cellcolor{shadow}33.0 & \cellcolor{shadow}84.9  \\
        & \cellcolor{shadow}\textsc{PMI}-3 & \cellcolor{shadow}\textbf{27.6} & \cellcolor{shadow}\textbf{83.5} & \cellcolor{shadow}\textbf{19.6} & \cellcolor{shadow}\textbf{80.7} & \cellcolor{shadow}\textbf{32.4} & \cellcolor{shadow}\textbf{80.4} & \cellcolor{shadow}16.0 & \cellcolor{shadow}74.4 & \cellcolor{shadow}27.5 & \cellcolor{shadow}82.9 & \cellcolor{shadow}33.8 & \cellcolor{shadow}\textbf{85.4}  \\
        & \cellcolor{shadow}\textsc{PMI}-6 & \cellcolor{shadow}26.8 & \cellcolor{shadow}83.3 & \cellcolor{shadow}19.5 & \cellcolor{shadow}80.5 & \cellcolor{shadow}32.2 & \cellcolor{shadow}\textbf{80.4} & \cellcolor{shadow}15.5 & \cellcolor{shadow}74.5 & \cellcolor{shadow}\textbf{28.5} & \cellcolor{shadow}\textbf{83.3} & \cellcolor{shadow}\textbf{33.9} & \cellcolor{shadow}85.3  \\

         \midrule
        \multirow{5}{*}{\textbf{Bloomz-176B$^\ast$}} & Direct & 24.0 & 78.4 & 16.0 & 76.4 & 27.3 & 77.1 & 13.0 & 70.7 & 29.5 & 83.9 & \ \ 5.6 & 53.8  \\
         & Pivot & 25.0 & 82.8 & 20.8 & 81.3 & 34.6 & 82.1 & \ \ 9.5 & 66.2 & 27.6 & 82.6 & 31.5 & \textbf{84.6}  \\
        & \cellcolor{shadow}\textsc{PMI}-1 & \cellcolor{shadow}25.4 & \cellcolor{shadow}80.7 & \cellcolor{shadow}17.3 & \cellcolor{shadow}77.6 & \cellcolor{shadow}33.1 & \cellcolor{shadow}80.4 & \cellcolor{shadow}11.9 & \cellcolor{shadow}70.0 & \cellcolor{shadow}28.0 & \cellcolor{shadow}82.4 & \cellcolor{shadow}23.5 & \cellcolor{shadow}75.8  \\
        & \cellcolor{shadow}\textsc{PMI}-3 & \cellcolor{shadow}28.2 & \cellcolor{shadow}\textbf{83.9} & \cellcolor{shadow}21.1 & \cellcolor{shadow}81.2 & \cellcolor{shadow}\textbf{35.7} & \cellcolor{shadow}\textbf{82.2} & \cellcolor{shadow}\textbf{16.0} & \cellcolor{shadow}\textbf{73.9} & \cellcolor{shadow}31.7 & \cellcolor{shadow}83.8 & \cellcolor{shadow}31.8 & \cellcolor{shadow}83.7  \\
        & \cellcolor{shadow}\textsc{PMI}-6 & \cellcolor{shadow}\textbf{28.3} & \cellcolor{shadow}83.8 & \cellcolor{shadow}\textbf{21.7} & \cellcolor{shadow}\textbf{81.4} & \cellcolor{shadow}36.6 & \cellcolor{shadow}82.9 & \cellcolor{shadow}15.0 & \cellcolor{shadow}73.5 & \cellcolor{shadow}\textbf{32.4} & \cellcolor{shadow}\textbf{84.7} & \cellcolor{shadow}\textbf{34.0} & \cellcolor{shadow}84.2  \\

        \bottomrule
    \end{tabular} 
    }

    \caption{Experiments on the WMT dataset. Note that the pivot row displays the maximum scores among all pivot prompts, and the order of the parallel languages indicates the priority when being integrated into \textsc{PMI}-$k$ prompts. $\dag$ and $\ast$ represent the model is fine-tuned or not respectively.}

    \label{tab:t4}
    

\end{table*}

\paragraph{Multilingual input inhibits neurons, whereas monolingual input activates neurons.} Figure \ref{fig:f4} (c) and (d) show the proportion of activated neurons caused by monolingual and multilingual input. We see that, compared to direct translation, though monolingual and multilingual input can achieve better performance, their influence on neurons is the opposite, i.e., monolingual input activates neurons, whereas multilingual input inhibits neurons. Moreover, PMI$_{GT}$ inhibits more neurons than PMI$_{ML}$ and PMI$_{MS}$ activates more neurons than PMI$_{PA}$.

\paragraph{PMI simulates a \textit{one-off} synaptic pruning.} During the maturation of biological brains, synaptic pruning is a necessary process that removes less commonly used neural connections, thus making frequently-used neural pathways more powerful and efficient \cite{huttenlocher1979synaptic,huttenlocher1990morphometric}. In other words, the brain benefits from little and precise neuron activation. We show that PMI simulates the synaptic pruning during the inference from two aspects: (1) as demonstrated above, PMI \textit{inhibits neurons}; (2) PMI \textit{promotes more precise neuron activation.} Figure \ref{fig:f5} records the activation state of the most commonly used neurons. It shows that compared to the baseline prompt, PMI promotes the activation of the top 1\% of neurons commonly used. Meanwhile, other neurons rarely used are activated fewer times to achieve an overall effect of inhibition, as shown in Figure \ref{app fig: activated neurons in Bloomz-176B on RTE}. This indicates that more targeted and effective neuron activation patterns—where some important neurons are activated more while others are less often—could be facilitated by PMI. Synaptic pruning occurs during the maturation of the brain, while PMI enhances models specifically at their inference stages, not during training. Therefore, we propose that PMI simulates a \textit{one-off} synaptic pruning, exerting a short-term effect on models.

\section{Wide Evaluation of PMI Without Human Translations}
\label{sec:s4}
Next, we focus on evaluating the PMI method on downstream tasks under real scenario setups.

\subsection{Tasks and Evaluation}

We totally evaluated PMI on six tasks. \textbf{(1) Machine Translation:} We conducted experiments on five high-resource directions of WMT22 and one low-resource direction of WMT21. \textbf{(2) Nature Language Inference:} We chose RTE \cite{DBLP:conf/nips/WangPNSMHLB19} and three languages in XNLI \cite{DBLP:conf/emnlp/ConneauRLWBSS18}. The metric was accuracy. \textbf{(3) Reading Comprehension:} We did evaluation on this long sequence task using BoolQ\footnote{This dataset is also leaked to Bloomz-176B.} \cite{DBLP:conf/naacl/ClarkLCK0T19}. Our metric was accuracy. \textbf{(4) Text Simplification:} We used Wiki-auto \cite{DBLP:conf/acl/JiangMLZX20}, and SARI\footnote{\url{https://github.com/feralvam/easse}} \cite{DBLP:conf/acl/Alva-ManchegoMB20} was chosen as the metric. \textbf{(5) Abstractive Summarization:} For this paragraph-level task, we mainly reported the performance on two languages in XLSum \cite{DBLP:conf/acl/HasanBIMLKRS21}. The metric was F1-Rouge\footnote{\url{https://github.com/Isaac-JL-Chen/rouge\_chinese}} \cite{lin-2004-rouge}. \textbf{(6) Mathematical Reasoning:} We conducted experiments on GSM8K \cite{DBLP:journals/corr/abs-2110-14168}. We also apply the chain-of-thought (CoT) technique \cite{DBLP:conf/nips/Wei0SBIXCLZ22} to explore whether PMI could enhance the reasoning capabilities of large language models (LLMs). The metric was accuracy. To streamline computation, we reconstructed our test set by randomly selecting 1000 samples from BoolQ, Wiki-auto, and XLSum, along with 3000 samples from XNLI, leaving other tasks unchanged.

\subsection{Models}
The experiment was conducted on 8 instruction-tuned open source multilingual LLMs whose parameters range from 7B to 176B, including LLaMA3-8B \cite{llama3modelcard}, Bloomz-176B \cite{DBLP:conf/acl/MuennighoffWSRB23}, Qwen-7B, -14B, -72B \cite{DBLP:journals/corr/abs-2309-16609}, ALMA-13B \cite{DBLP:journals/corr/abs-2309-11674}, Yi-34B \cite{Yi} and mT0-13B \cite{DBLP:journals/corr/abs-2211-05100}. We also evaluated the effectiveness of PMI on two commercial ones, involving ChatGPT and GPT-4. All of them are pre-trained with multilingual corpora except for ALMA-13B, which is specially fine-tuned for the MT task based on LLaMA2-13B \cite{DBLP:journals/corr/abs-2307-09288}. Other details about models, training, and decoding setups can be found in Appendix \ref{app sec: Details of Experiment Setups}.

\subsection{Baselines}

\paragraph{Direct Prompt} means that, given the original input, LLMs accomplish the task directly. Here, we report the results of one-shot on ChatGPT while zero-shot on others for the best performance.

\paragraph{Pivot Prompt} indicates that the original input is translated into a parallel language, and LLMs are fed with the translation to accomplish the task. To ensure high-quality translations and the reproducibility of our study, we utilized the publicly and easily accessible GPT-4 for translating the WMT and GSM8K datasets. For other datasets, we employed ChatGPT. We display the maximum scores of pivot prompts; see Appendix \ref{app sec: Full Experimental Results of Pivot Prompts} for full results.

\subsection{Results and Analyses}

\paragraph{PMI effectively pushes the boundaries across various tasks and languages.} Table \ref{tab:t4} suggests that PMI achieves superior results across 6 translation directions, including high-resource and low-resource source languages. Additionally, Tables \ref{tab:t5} and \ref{tab:t6} show PMI's competitive edge against baselines in various tasks, irrespective of text length. Furthermore, in Table \ref{app tab: Comparing the performance between few-shot and PiM}, we can see that PMI outperforms few-shot learning on the translation task, especially in terms of the COMET score.

We also evaluate the effectiveness of PMI on mathematical reasoning tasks and CoT scenarios. Table \ref{tab:reasoning} suggests that PMI can further boost the superior reasoning performance of GPT models, with accuracy nearly reaching 96\% on the GSM8K benchmark. Beyond the noted improvements in the commonly used 5-shot and 8-shot scenarios, we also observed significant performance gains with PMI in 0-shot settings for GPT-4. We attribute this to PMI aiding LLMs in gaining a more comprehensive understanding of the tasks in scarce-shot scenarios.

\paragraph{Weak model augments strong model.} Table \ref{tab:Experiments of GPT-4o on WMT} shows that when we utilize parallel multilingual translations from GPT-4 to augment a stronger LLM like GPT-4o, the performance of GPT-4o+PMI surpasses two exceptional baselines, including GPT-4 and GPT-4o. It underscores the necessity of using PMI instead of relying solely on a remarkable MT system. Also, this demonstrates that PMI still yields better performance when the parallel translations come from a weak model, further validating its effectiveness and practicality.

\begin{table}[t!]
\LARGE
    \centering

 \resizebox{1.0\linewidth}{!}{
    \begin{tabular}{ll|c|ccc|c}
        \toprule
        \multicolumn{2}{c|}{\multirow{2.5}{*}{\textbf{System}}}& \multicolumn{5}{c}{\textbf{Accuracy}} \\ 
        \cmidrule(l){3-7}
        & & \textbf{RTE} & \multicolumn{3}{|c|}{\textbf{XNLI}} & \textbf{BoolQ} \\
        \midrule
        \multicolumn{2}{c|}{\textbf{\textit{Source Language}}} & \textit{En} & \textit{Fr} & \textit{De} & \textit{Zh} & \textit{En}\\
        \midrule
        \multicolumn{2}{c|}{\textbf{\textit{Parallel Languages}}} & \textit{Es Fr De} & \textit{Es Ru De} & \textit{Es Ru Fr} & \textit{Es Fr De} & \textit{Es} \\
        
        \midrule
        \multirow{3}{*}{\textbf{Qwen-7B$^\dag$}} & Direct & 91.3 & 79.9 & 76.7 & 78.2 & 86.0  \\
        & Pivot & 86.6 & 78.9 & 80.2 & 74.2 & 83.3  \\
        & \cellcolor{shadow}\textsc{PMI} & \cellcolor{shadow}\textbf{91.7} & \cellcolor{shadow}\textbf{80.7} & \cellcolor{shadow}\textbf{80.6} & \cellcolor{shadow}\textbf{80.7} & \cellcolor{shadow}\textbf{86.7}  \\

        \midrule
        \multirow{3}{*}{\textbf{Qwen-14B$^\dag$}} & Direct & 91.3 & 81.5 & 78.2 & 80.6 & 88.5  \\
        & Pivot & 90.6 & 80.5 & 79.8 & 74.2 & 86.0  \\
        & \cellcolor{shadow}\textsc{PMI} & \cellcolor{shadow}\textbf{92.4} & \cellcolor{shadow}\textbf{81.6} & \cellcolor{shadow}\textbf{80.7} & \cellcolor{shadow}\textbf{80.7} & \cellcolor{shadow}\textbf{89.0}  \\

        \midrule
        \multirow{3}{*}{\textbf{Qwen-72B$^\dag$}} & Direct & 91.7 & \textbf{86.4} & 84.4 & \textbf{84.6} & 91.2  \\
        & Pivot & 92.4 & 85.8 & 85.5 & 80.6 & 89.1  \\
        & \cellcolor{shadow}\textsc{PMI} & \cellcolor{shadow}\textbf{92.4} & \cellcolor{shadow}\textbf{86.4} & \cellcolor{shadow}\textbf{85.6} & \cellcolor{shadow}\textbf{84.6} & \cellcolor{shadow}\textbf{91.9}  \\

        \midrule
        \multirow{3}{*}{\textbf{ALMA-13B$^\dag$}} & Direct & 89.5 & 82.1 & 79.3 & 77.5 & 86.5  \\
        & Pivot & 84.5 & 82.0 & 80.8 & 75.9 & 81.1  \\
        & \cellcolor{shadow}\textsc{PMI} & \cellcolor{shadow}\textbf{90.3} & \cellcolor{shadow}\textbf{83.8} & \cellcolor{shadow}\textbf{81.9} & \cellcolor{shadow}\textbf{78.8} & \cellcolor{shadow}\textbf{87.4}  \\

        \midrule
        \multirow{3}{*}{\textbf{Yi-34B$^\dag$}} & Direct & 92.1 & 70.0 & 66.8 & 72.0 & 89.6  \\
        & Pivot & 85.9 & 71.5 & 72.6 & 68.1 & 86.8  \\
        & \cellcolor{shadow}\textsc{PMI} & \cellcolor{shadow}\textbf{93.1} & \cellcolor{shadow}\textbf{73.1} & \cellcolor{shadow}\textbf{73.7} & \cellcolor{shadow}\textbf{72.6} & \cellcolor{shadow}\textbf{90.2}  \\

        \midrule
        \multirow{3}{*}{\textbf{Bloomz-176B$^\ast$}} & Direct & 76.5 & 53.9 & 50.5 & 53.9 & -  \\
        & Pivot & 77.6 & 53.1 & \textbf{53.3} & 53.7 & -  \\
        & \cellcolor{shadow}\textsc{PMI} & \cellcolor{shadow}\textbf{82.0} & \cellcolor{shadow}\textbf{57.3} & \cellcolor{shadow}52.5 & \cellcolor{shadow}\textbf{54.9} & \cellcolor{shadow}-  \\

        \bottomrule
    \end{tabular} 
    }

    \caption{Experiments on NLU tasks. We apply \textsc{PMI}-3 across all tasks, with the exception of the reading comprehension task, for which we apply \textsc{PMI}-1.}

    \label{tab:t5}
    

\end{table}

\paragraph{Automatic translation triggers learning from PMI.} Since the lack of high-quality human translation, all the translations used in experiments come from GPT-4 or ChatGPT. We see, on the one hand, PMI powered by MT outperforms pivot prompts. Even though some pivot prompts perform worse than the direct prompt, integrating these languages into PMI still enhances the comprehension of LLMs. On the other hand, Figure \ref{app fig: The translation performance and the activation proportion of different prompts on WMT dataset} shows that PMI armed with MT achieves improvements by inhibiting neurons and promoting more precise activation. These results demonstrate the consistent learning behavior triggered by translations from human experts and MT systems.

\paragraph{Few-shot learning performs similarly to PMI.} Table \ref{tab:t7} and Figure \ref{app fig: activated neurons in Bloomz-176B on RTE} suggest that few-shot learning also inhibits neurons and facilitates more precise activation, and combining few-shot learning and PMI further enhances this neuron reaction.

\begin{table}[t!]
\LARGE
    \centering

 \resizebox{0.85\linewidth}{!}{
    \begin{tabular}{ll|c|c|c}
        \toprule
        \multicolumn{2}{c|}{\multirow{2.5}{*}{\textbf{System}}} & \textbf{SARI} & \multicolumn{2}{c}{\textbf{R2 / RL}} \\
        
        \cmidrule(l){3-5}
         & & \textbf{Wiki-Auto} & \multicolumn{2}{c}{\textbf{XLSum}} \\

        \midrule
        \multicolumn{2}{c|}{\textbf{\textit{Source Language}}} & \textit{En} & \textit{Es} & \textit{Ru} \\

        \midrule
        \multicolumn{2}{c|}{\textbf{\textit{Parallel Languages}}} & \textit{Es Fr De} & \textit{Fr} & \textit{Es} \\
        
        \midrule
        \multirow{3}{*}{\textbf{Qwen-7B$^\dag$}} & Direct & 45.6 & 10.7 / 23.5 & \textbf{45.4 / 41.6}  \\
         & Pivot & 43.2 & \ \ 9.4 / 22.7 & 41.1 / 38.6  \\
         & \cellcolor{shadow}\textsc{PMI} & \cellcolor{shadow}\textbf{47.6} & \cellcolor{shadow}\textbf{11.0 / 23.6} & \cellcolor{shadow}45.3 / 41.1  \\
         
        \midrule
        \multirow{3}{*}{\textbf{Qwen-14B$^\dag$}} & Direct & 46.2 & 12.2 / 24.7 & 46.6 / 42.7  \\
         & Pivot & 43.8 & \ \ 9.0 / 21.4 & 40.2 / 38.3  \\
         & \cellcolor{shadow}\textsc{PMI} & \cellcolor{shadow}\textbf{48.9} & \cellcolor{shadow}\textbf{12.7 / 25.4} & \cellcolor{shadow}\textbf{47.9 / 43.1}  \\

        \midrule
        \multirow{3}{*}{\textbf{ALMA-13B$^\dag$}} & Direct & 45.7 & \textbf{12.1 / 24.8} & 47.7 / 43.5  \\
         & Pivot & 43.2 & 10.4 / 22.9 & 44.3 / 41.2  \\
         & \cellcolor{shadow}\textsc{PMI} & \cellcolor{shadow}\textbf{47.5} & \cellcolor{shadow}11.5 / 24.5 & \cellcolor{shadow}\textbf{47.7 / 43.9}  \\

        \midrule
        \multirow{3}{*}{\textbf{Yi-34B$^\dag$}} & Direct & 45.4 & 11.8 / 24.6 & 45.4 / 41.5  \\
         & Pivot & 43.5 & 10.6 / 23.3 & 41.7 / 38.8  \\
         & \cellcolor{shadow}\textsc{PMI} & \cellcolor{shadow}\textbf{47.2} & \cellcolor{shadow}\textbf{12.0 / 24.6} & \cellcolor{shadow}\textbf{45.5 / 41.8}  \\  

        \bottomrule
    \end{tabular} 
    }

    \caption{Experiments on other NLG tasks. We employ \textsc{PMI}-3 and \textsc{PMI}-1 for the text simplification and abstractive summarization task respectively.}

    \label{tab:t6}
    

\end{table}

\begin{table}[t!]
\Large
    \centering

 \resizebox{0.8\linewidth}{!}{
    \begin{tabular}{lllll}
        \toprule
        \multirow{2.5}{*}{\textbf{System}} & & \multicolumn{3}{c}{\textbf{GSM8K CoT}}  \\ 
        \cmidrule(lr){3-5}
        & & \textbf{0-shot} & \textbf{5-shot} & \textbf{8-shot} \\
         
        \midrule
        \multirow{3}{*}{\textbf{GPT-4o}} & Direct & 86.9 & 94.5 & 94.9 \\
         & \cellcolor{shadow}PMI-3 & \cellcolor{shadow}86.5$^{\downarrow 0.4}$ & \cellcolor{shadow}95.1$^{\uparrow 0.6}$ & \cellcolor{shadow}95.2$^{\uparrow 0.3}$ \\
         & \cellcolor{shadow}PMI-6 & \cellcolor{shadow}\textbf{87.0}$^{\uparrow 0.1}$ & \cellcolor{shadow}\textbf{95.2}$^{\uparrow 0.7}$ & \cellcolor{shadow}\textbf{95.9}$^{\uparrow 1.0}$ \\
         
        \midrule
        \multirow{3}{*}{\textbf{GPT-4}} & Direct & 64.6 & 92.8 & 93.3 \\
         & \cellcolor{shadow}PMI-3 & \cellcolor{shadow}74.7$^{\uparrow 10.1}$ & \cellcolor{shadow}\textbf{93.3}$^{\uparrow 0.5}$ & \cellcolor{shadow}93.3$^{\uparrow 0.0}$ \\
         & \cellcolor{shadow}PMI-6 & \cellcolor{shadow}\textbf{76.2}$^{\uparrow 11.6}$ & \cellcolor{shadow}\textbf{93.3}$^{\uparrow 0.5}$ & \cellcolor{shadow}\textbf{93.7}$^{\uparrow 0.4}$ \\

        \bottomrule
    \end{tabular} 
    }

    \caption{Experiments on the mathematical reasoning.}

    \label{tab:reasoning}
    
\vspace{-1.1em}

\end{table}


\paragraph{Superiority of PMI remains when English is the original or parallel language.} Despite the subtle improvements on FLORES-200 En $\rightarrow$ De in Section \ref{sec:s1}, results of RTE, BoolQ, and WMT De $\rightarrow$ Fr show that PMI not only achieves prime performance on English-source inputs but also outperforms all pivot prompts when we choose English as one of the parallel languages.

We discuss the fine-tuning demands of PMI in Appendix \ref{app sec: Effectiveness of PMI on more modern LLMs}, self-augmentation in Appendix \ref{app sec: Self-augmentation}, and the trade-off between the inference speed and improvements in Appendix \ref{app sec: Inference Speed}.

\section{Related Work}

\paragraph{Multi-way Neural Machine Translation.}
Multi-way input is a successful method to enhance multilingual neural machine translation (MNMT) systems by providing the source language and its translations in different languages \cite{DBLP:conf/mtsummit/OchN01}. In the inference stage, most works rely on high-quality translations from human experts \cite{DBLP:conf/naacl/ZophK16,DBLP:conf/emnlp/FiratSAYC16,DBLP:conf/iwslt/NishimuraSN018,DBLP:conf/lrec/ChoiSK18}. However, this ground truth multilingual data is scarce in reality, limiting the application of multi-way input. Different from multi-way MNMT, we find that LLMs benefit from PMI even when parallel multilingual input is derived from automatic MT systems, enabling us to apply PMI on a wide range of tasks.

\paragraph{Statistics of Activated Neurons in Transformer Models.}
Recently, statistics of activated neurons in transformer models by counting nonzero values in the output of ReLU is introduced by \citet{DBLP:conf/acl/ZhangL00S022}. Moreover, \citet{DBLP:conf/iclr/LiYBLRRYCYGK23} show that the sparse activation of neurons is ubiquitous. In this work, we extend the statistical method to advanced transformer architectures. We hope this effort can help deepen our insights into the learning mechanism behind LLMs.

\begin{table}[t!]
\Large
    \centering

 \resizebox{0.85\linewidth}{!}{
    \begin{tabular}{ll|cc|cc}
        \toprule
        \multicolumn{2}{c|}{\textbf{System}} & \textbf{BLEU} & \textbf{COMET} & \textbf{BLEU} & \textbf{COMET} \\
        \midrule
        \multicolumn{2}{c|}{\textbf{\textit{Direction}}} & \multicolumn{2}{c|}{\textit{De $\rightarrow$ Fr}}  & \multicolumn{2}{c}{\textit{Zh $\rightarrow$ En}}  \\

        \midrule
        \multicolumn{2}{l|}{\textbf{GPT-4}} & 39.0 & 84.3 & 23.2 & 81.6  \\
        \midrule
        \multirow{2}{*}{\textbf{GPT-4o}} & Direct & 39.2 & 83.1 & 23.1 & \textbf{82.4}  \\
         & \cellcolor{shadow}PMI & \cellcolor{shadow}\textbf{42.5} & \cellcolor{shadow}\textbf{84.8} & \cellcolor{shadow}\textbf{23.6} & \cellcolor{shadow}\textbf{82.4} \\

        \midrule
        \multicolumn{2}{c|}{\textbf{\textit{Direction}}} & \multicolumn{2}{c|}{\textit{En $\rightarrow$ De}}  & \multicolumn{2}{c}{\textit{En $\rightarrow$ Zh}}  \\

        \midrule
        \multicolumn{2}{l|}{\textbf{GPT-4}} & 35.5 & 87.2 & 42.5 & 86.4  \\
        \midrule
        \multirow{2}{*}{\textbf{GPT-4o}} & Direct & \textbf{36.8} & 87.5 & \textbf{44.5} & 87.6  \\
         & \cellcolor{shadow}PMI & \cellcolor{shadow}36.3 & \cellcolor{shadow}\textbf{88.0} & \cellcolor{shadow}\textbf{45.5} & \cellcolor{shadow}\textbf{87.7} \\

        \bottomrule
    \end{tabular} 
    }

    \caption{Experiments of GPT-4o on WMT. We report the best performance among PMI-1, PMI-3, and PMI-6 in the PMI lines.}

    \label{tab:Experiments of GPT-4o on WMT}
    
\vspace{-0.7em}

\end{table}

\paragraph{Cross-lingual In-context Learning.}
Several works have investigated cross-lingual prompts \cite{DBLP:journals/corr/abs-2302-14229,DBLP:conf/iclr/ShiSF0SVCTRZ0W23,DBLP:conf/acl/MuRCFLLXZZ23}. One line of research requests LLMs to address the input problem in multiple languages orderly, then emphasizes self-consistency by aligning results of these languages to improve performance on reasoning tasks \cite{DBLP:conf/emnlp/0001CWHC23}. To augment LLMs' performance with multilingual input, other works encourage LLMs to rephrase the input in English and then perform step-by-step analysis, indeed turning English into a pivot language \cite{DBLP:conf/emnlp/HuangTZZSXW23,DBLP:journals/corr/abs-2311-08711,DBLP:journals/corr/abs-2306-11372}. Our work, in contrast, explores the behavior of LLMs that learns from parallel input in multiple languages simultaneously, revealing a new ICL capability.

\begin{table}[t!]
\LARGE
    \centering

 \resizebox{1.0\linewidth}{!}{
    \begin{tabular}{cccc|cccc}
        \toprule
        \multicolumn{4}{c|}{\textbf{Qwen-14B}} & \multicolumn{4}{c}{\textbf{Bloomz-176B}} \\

        \midrule
        \multicolumn{2}{c}{\textbf{XNLI (De)}} & \multicolumn{2}{c|}{\textbf{Wiki-Auto}}  & \multicolumn{4}{c}{\textbf{RTE}} \\
        Direct & \textsc{PMI-3} & Direct & \textsc{PMI-3} & Direct & \textsc{PMI-3} & 5-shot & \makecell{5-shot \\ + \textsc{PMI-3}} \\
        
        \midrule
        \multicolumn{2}{c}{\textit{Accuracy}} & \multicolumn{2}{c|}{\textit{SARI}} & \multicolumn{4}{c}{\textit{Accuracy}} \\
        78.2 & 80.7 & 46.2 & 49.0 & 76.5 & 82.0 & 80.1 & 81.2 \\

        \midrule
        \multicolumn{4}{c|}{\textit{Activation Proportion (\%)}} & \multicolumn{4}{c}{\textit{Activation Proportion (\%)}} \\
        
        29.5 & 29.3 & 28.7 & 28.4 & 4.4 & 4.3 & 4.1 & 3.9 \\
        \bottomrule
    \end{tabular} 
    }

    \caption{The performance and activation proportion of conventional ICL and \textsc{PMI} on NLU and NLG tasks.}
    \label{tab:t7}
    

\end{table}

\section{Conclusions}
We reveal that LLMs can learn from parallel multilingual input. Firstly, comprehensive experiments across 8 typical datasets, 10 commonly used multilingual LLMs, and 7 languages demonstrate the effectiveness and applicability of PMI. Secondly, statistics of activated neurons indicate that PMI optimizes performance by inhibiting neurons and promoting more precise neuron activation, which performs like one-off synaptic pruning.  In future work, we aim to explore applying PMI to multimodal tasks and observing neural activation behaviors in large multimodal models.

\section{Limitations}
In fact, during the inference, LLMs will inevitably refer to the semantics of the translation in PMI to understand the input comprehensively. As a result, though our extensive experiments have demonstrated that LLMs can benefit from PMI, the quality of translation will influence the final performance. On the other hand, we do not discuss the effect of cross-language, such as code-switch multilingual prompts, because it deviates from the intention of PMI, i.e., providing parallel input. However, it is still a potential direction, and we leave it for future work.

\section*{Acknowledgements}
This work was supported in part by the National Science Foundation of China (No.62276056), the Natural Science Foundation of Liaoning Province of China (2022-KF-16-01), the Fundamental Research Funds for the Central Universities (Nos. N2216016 and N2316002), the Yunnan Fundamental Research Projects (No. 202401BC070021), and the Program of Introducing Talents of Discipline to Universities, Plan 111 (No.B16009). The authors would like to thank Yunhe Gao, Chi Hu, Erfeng He, and anonymous reviewers for their advice.

\bibliography{custom}

\appendix
\label{sec:appendix}

\begin{figure}
\centering
\includegraphics[width=0.47\textwidth]{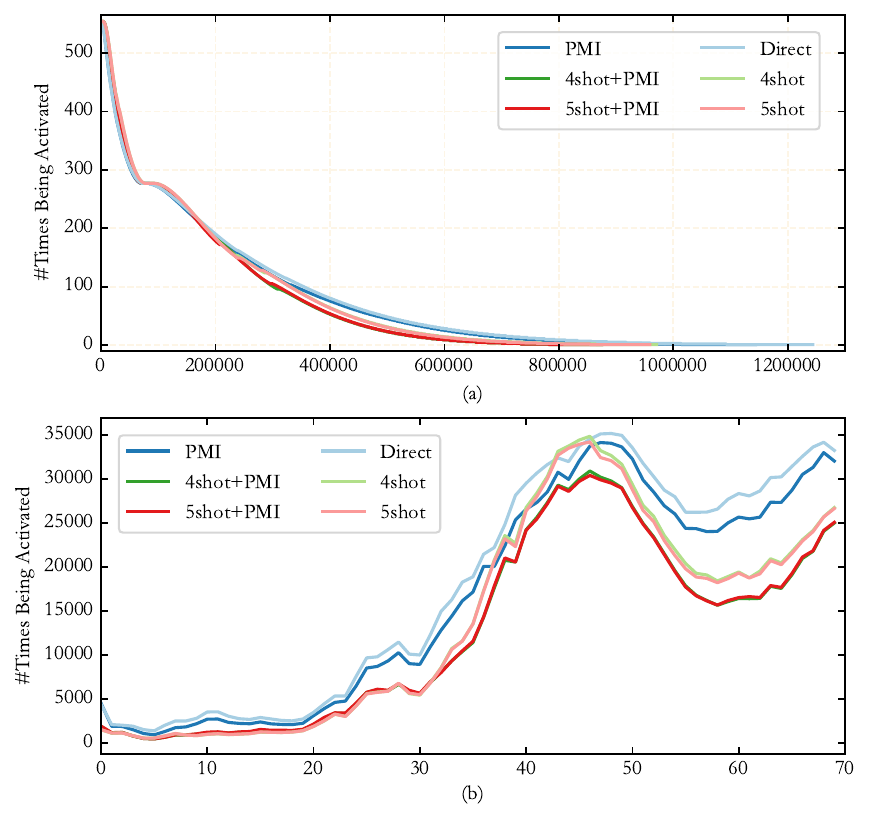}
\caption{Distribution of all activated neurons in Bloomz-176B on RTE. The horizontal axis of the figure (a) represents different neurons arranged in descending order of the number of times being activated, and the horizontal axis of the figure (b) stands for the number of transformer layers.}
\label{app fig: activated neurons in Bloomz-176B on RTE}
\end{figure}

\begin{figure}
\centering
\includegraphics[width=0.47\textwidth]{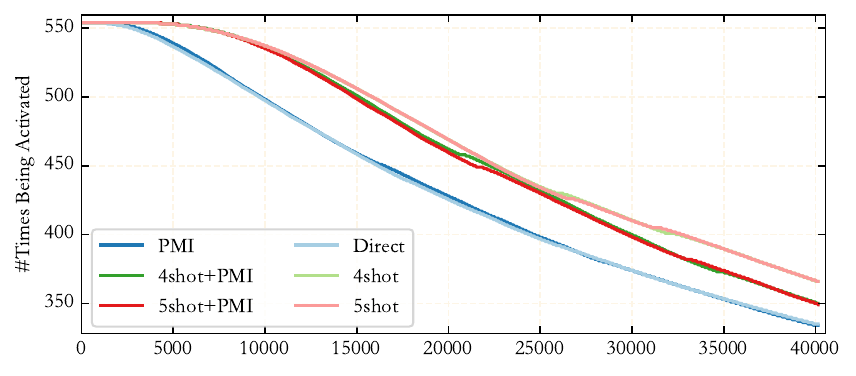}
\caption{The distribution of the top 1\% of activated neurons in Bloomz-176B on RTE.}
\label{app fig: top 0.01 of activated neurons in Bloomz-176B on RTE}
\end{figure}

\begin{figure}
\centering
\includegraphics[width=0.48\textwidth]{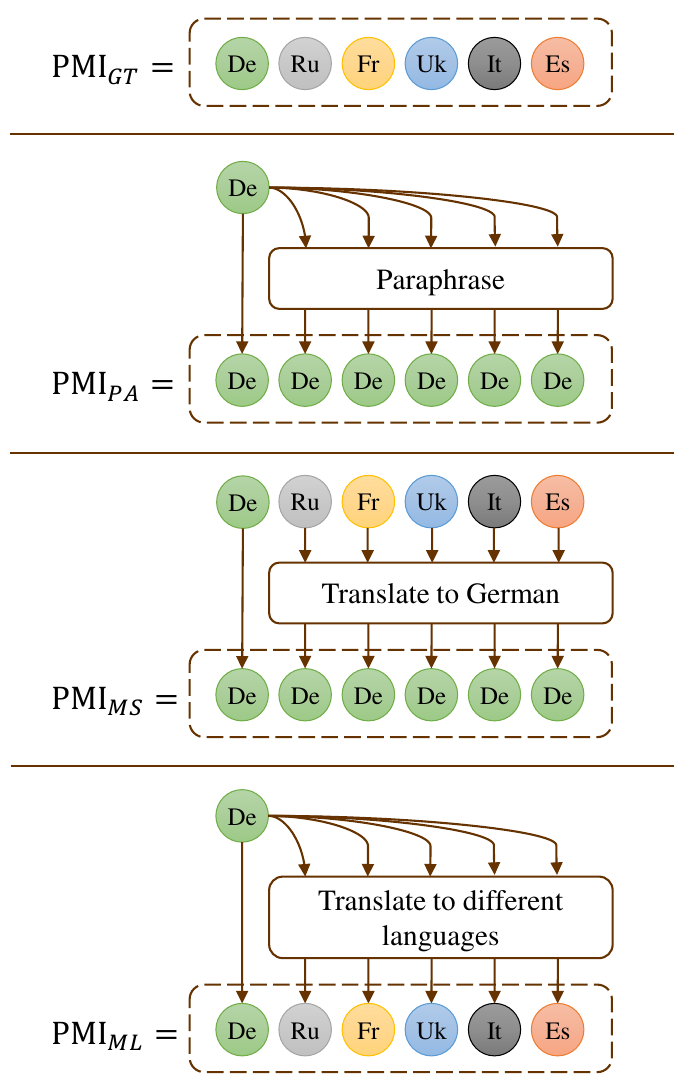}
\vspace{-1.5em}
\caption{An illustration of different strategies for constructing parallel inputs in Section \ref{sec:s2}. Taking De $\to$ En translation as an example, $\textsc{PMI}_{GT}$ consists of multilingual human translations from several experts; $\textsc{PMI}_{PA}$ is made up of monolingual sentences paraphrased from the original German input; $\textsc{PMI}_{MS}$ is composed of German translations where their source language texts are from different experts; and $\textsc{PMI}_{ML}$ includes multilingual translations of the original German input derived from a single translator.}
\label{fig:illustration of ablation study prompts}
\vspace{-0.7em}
\end{figure}

\section{Design of Prompts}
To prohibit LLMs from skewing towards any particular languages in the input, we don't point out the original input of tasks in our prompts. All of the prompts are listed in Table \ref{app tab: All the prompts used in experiments}. In this table, the content that is italicized and highlighted in gray indicates variable elements, which should be replaced according to the specific task requirements.

\section{More Details About Statistical Method of Activated Neurons}
\label{app sec: More explanations of counting activated neurons}

\paragraph{Implementation of Counting Activated Neurons.} During the inference stage, each time LLMs calculate the representation of a token including input and output, the intermediate result of MLPs stands for an activation state of neurons. It is essential to note that \textit{we only make statistics of activated neurons based on the intermediate result corresponding to the output tokens.} This implementation is based on two concerns: (1) only the activation state of neurons corresponding to the output tokens directly contributes to the model-generated results. (2) since different prompting strategies differ in the length of input significantly, if the statistics are made based on both input and output tokens, then the results will be disturbed by the factor of length but not the actual impact of prompts, resulting in misdirected conclusions.

\paragraph{Activation Functions Used in LLMs.}Table \ref{app tab: activation functions} records some popular LLMs and the activation functions they used. 

\section{Supplementary Results About Neuron Activation}

\begin{table*}[t!]
\small
    \centering

 \resizebox{0.8\linewidth}{!}{
    \begin{tabular}{lll}
        \toprule
        \textbf{Activation Function} &  \textbf{Formula} & \textbf{Model} \\

        \midrule
         ReLU & $\text{max}\left(x,0\right)$ & Vanilla Transformer \\
         GELU & $0.5x\left(1+\text{erf}\left(x/\sqrt{2}\right)\right)$ & Bloom, Falcon \\
         SiLU & $x/\left(1+e^{-x}\right)$ & \textbackslash \\
         GEGLU & $\text{GELU}\left(X W_{up}\right) \odot \left(X V_{up}\right) $ & mT0 \\
         SwiGLU & $\text{SiLU}\left(X W_{up}\right) \odot \left(X V_{up}\right) $ & LLaMA, Qwen, ALMA, Yi \\

        \bottomrule
    \end{tabular} 
    }

    \caption{The activation functions of some commonly used multilingual LLMs. In GELU, the $\text{erf}(\cdot)$ stands for the Gauss Error Function. Note that our extended statistical method can be applied to all LLMs shown in this table.}

    \label{app tab: activation functions}

\end{table*}

\begin{figure*}
\centering
\subcaptionbox{Qwen-14B}{\includegraphics[width = 0.48\textwidth]{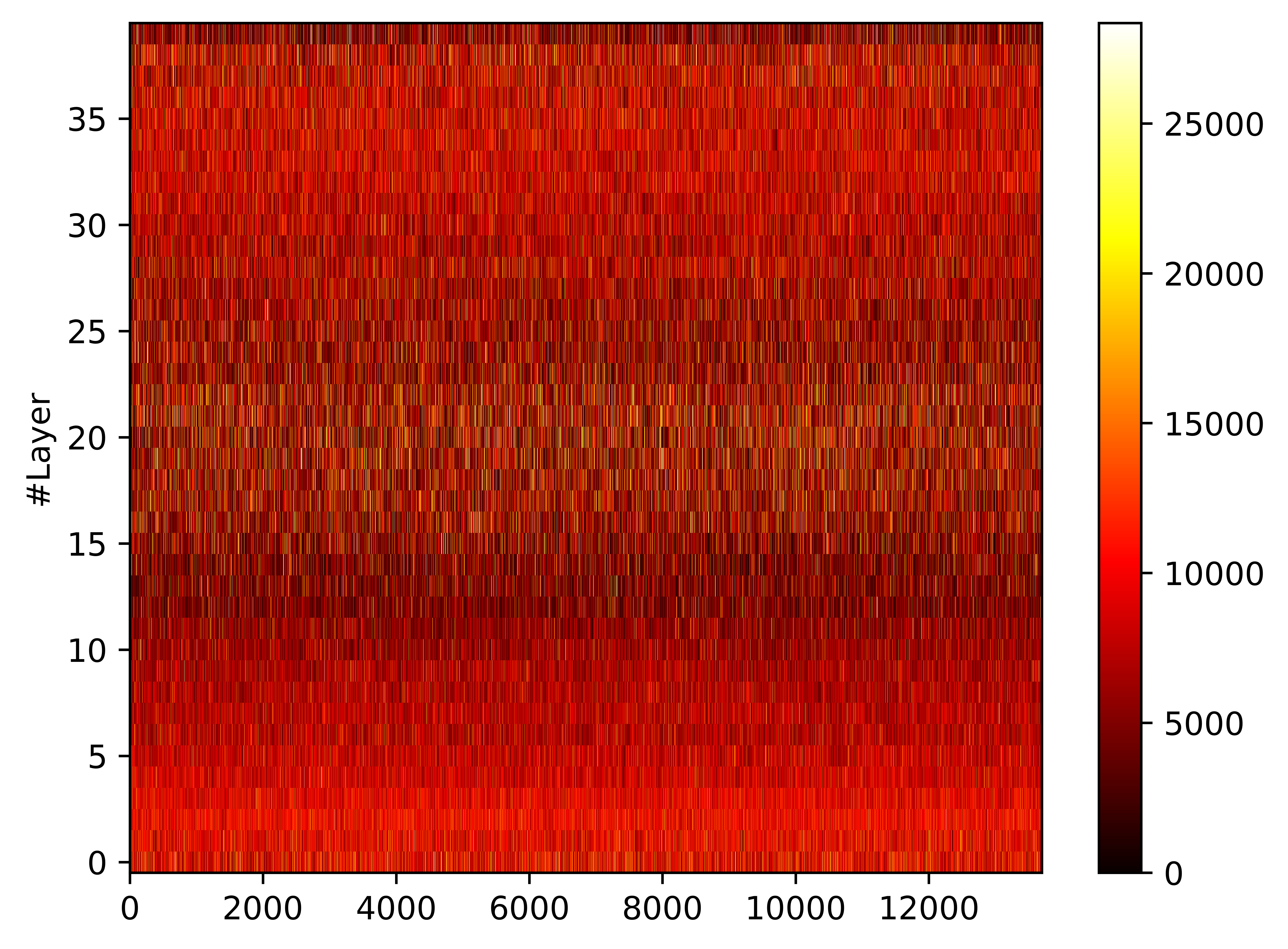}}
\hfill
\subcaptionbox{Bloomz-176B}{\includegraphics[width = 0.48\textwidth]{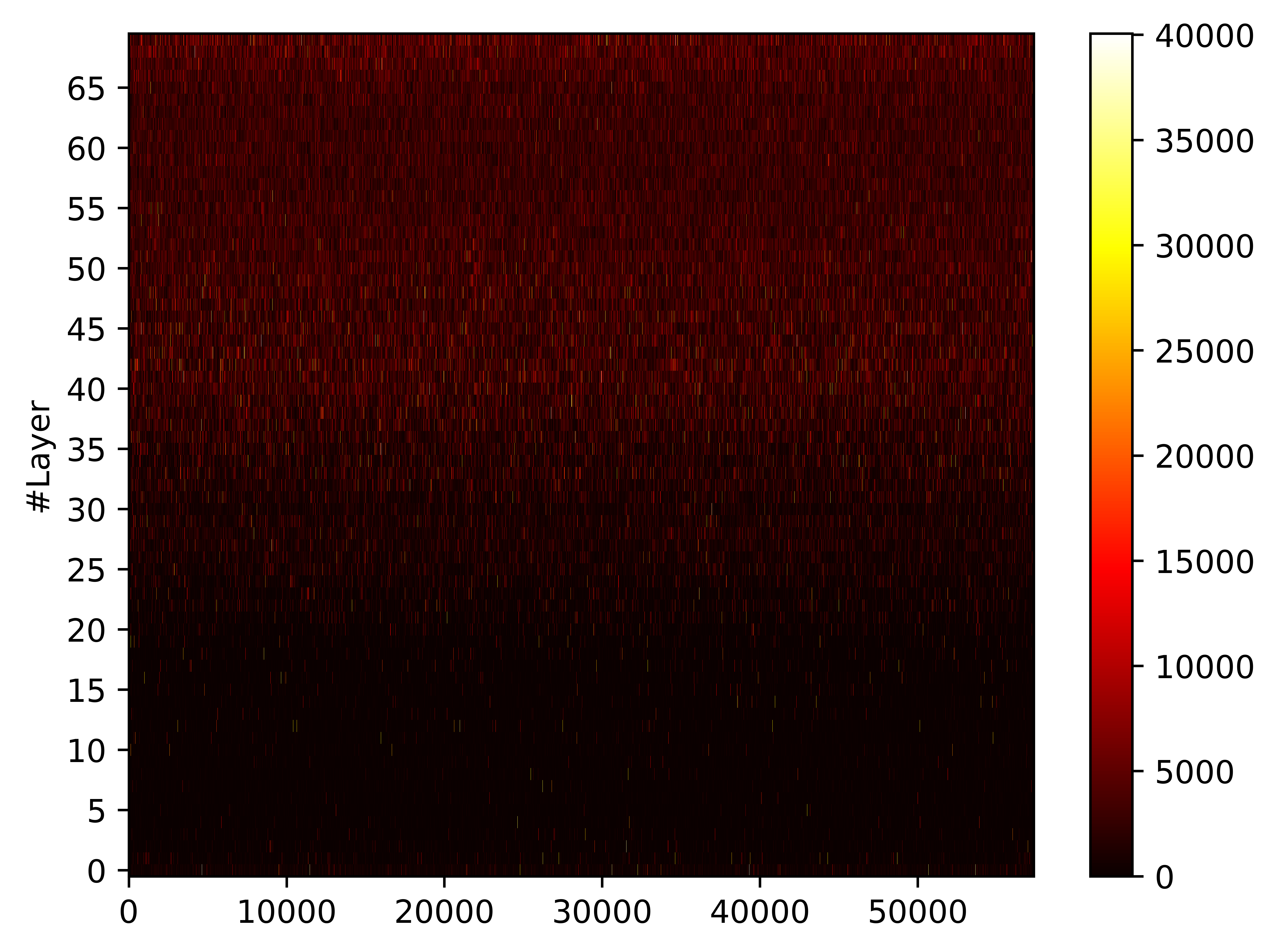}}
\caption{The heat maps of activated neurons in MLPs of Qwen-14B and Bloomz-176B when using the PMI-5 to translate De $\rightarrow$ En in the FLORES-200 and WMT dataset, respectively. The horizontal axis represents the dimension of the middle outputs in MLPs (i.e., each neuron). The vertical axis represents the number of layers in the model. And each element in the map stands for the number of times of was activated during the inference stage.}
\label{app fig: heatmap of activation neuron}
\vspace{-1.0em}
\end{figure*}

\begin{figure}
\centering
\includegraphics[width=0.47\textwidth]{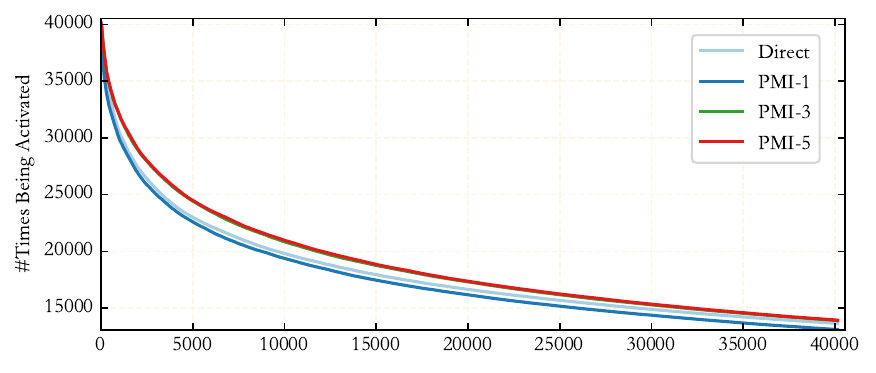}
\caption{The distribution of the top 1\% of activated neurons in Bloomz-176B on WMT22 De $\rightarrow$ En. The horizontal axis represents different neurons arranged in descending order of the number of times being activated.}
\label{app fig: Distribution of activated neurons in Bloomz-176B on WMT22}
\vspace{-0.7em}
\end{figure}

In Figure \ref{app fig: activated neurons in Bloomz-176B on RTE} (a), we can see that: (1) in the interval from 0 to 200000, the curves of PMI, few-shot learning and their combination are above that of baseline (i.e., Direct), indicating that they activate top 200,000 commonly used neurons; (2) beyond the 200,000 mark, these curves are below the curve of baseline, demonstrating that these prompts perform inhibiting other less used neurons. Furthermore, in Figure \ref{app fig: activated neurons in Bloomz-176B on RTE} (b), we can see that the inhibited neurons concentrate in the back two-thirds of the model layers. Figures \ref{app fig: Distribution of activated neurons in Bloomz-176B on WMT22} and \ref{app fig: top 0.01 of activated neurons in Bloomz-176B on RTE} report the distribution of the top 1\% of activated neurons in Bloomz-176B where PMI shows a clear impact of activation on most commonly used neurons. 

To visualize the activation happening in each neuron, in Figure \ref{app fig: heatmap of activation neuron}, we draw heat maps of Qwen-14B and Bloomz-176B when using the PMI-5 to translate De $\rightarrow$ En in the FLORES-200 and WMT datasets, respectively. It suggests that the neurons of Qwen-14B are more active while those of Bloomz-176B seem lazy and are activated fewer times. Furthermore, in each model, there are significant differences in the number of times being activated among different layers.

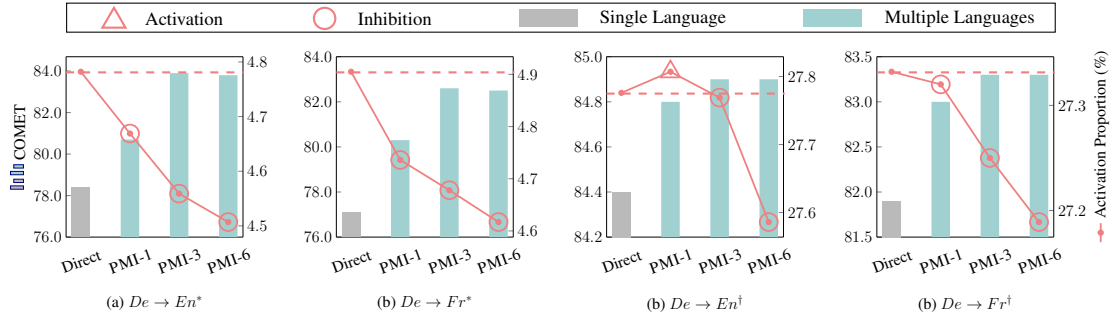
\begin{figure*}
\centering
\begin{tikzpicture}[scale=0.42]
\tikzstyle{every node}=[font=\Large]

    \node [rectangle,draw=black,minimum height=1.0em,minimum width=33.8em,font=\small,anchor=west,align=center] (final) at (0em,18em){};

    \node [anchor=north,align=center,minimum size=1pt,scale=0.50] (a) at (1.5,7.4){ \ref{pgfplots:triangle}};
    \node [anchor=center] at (3.7,6.9){\scriptsize Activation};

    \node [anchor=north,align=center,minimum size=1pt,scale=0.50] (a) at (8.1,7.4){ \ref{pgfplots:circle}};
    \node [anchor=center] at (10.3,6.9){\scriptsize Inhibition};  
    
    \node [rectangle,draw=none,fill=cGray,inner sep=2pt,minimum height=0.6em,minimum width=2em,font=\small,anchor=north,align=center] (a) at (15.1,7.2){};
    \node [anchor=center] at (18.7,6.9){\scriptsize Single Language};
    
    \node [rectangle,draw=none,fill=cGreen,inner sep=2pt,minimum height=0.6em,minimum width=2em,font=\small,anchor=north,align=center] (a) at (24,7.2){};
    \node [anchor=center] at (28,6.9){\scriptsize Multiple Languages};

    \begin{axis}[
      x=4em,
      at={(0,0)},
      ybar,
      bar width=.58cm,
      bar shift=0pt,
      axis y line*=left,
      ymin=76,
      xtick={1,2,3,4},
      x tick style={draw=none},
      xticklabels={Direct,\textsc{PMI}-1,\textsc{PMI}-3,\textsc{PMI}-6},
      xticklabel style={rotate=20, anchor=center, xshift=-0.5em, yshift=-1.5em},
      xlabel=(a) $De \rightarrow En$$^\ast$,
      x label style={yshift=-1.0em},
      ylabel=\ref{pgfplots:gray bar} \ref{pgfplots:green bar}  COMET,
      y tick label style={
        /pgf/number format/.cd,
            fixed,
            fixed zerofill,
            precision=1,
        /tikz/.cd
        },
    ]
        \addplot [
            draw = none,
            fill = cGray
        ]  coordinates {(1,78.4)};
        \addplot [
            draw = none,
            fill = cGreen
        ]  coordinates {(2,80.7) (3,83.9) (4,83.8)};
    \end{axis}
    
    \begin{axis}[
      x=4em,
      at={(0,0)},
      axis y line*=right,
      axis x line=none,
      xtick=data,
      xticklabel style={rotate=20, anchor=center, xshift=0em, yshift=-1.0em},
      symbolic x coords={Direct,\textsc{PMI}-1, \textsc{PMI}-3, \textsc{PMI}-6},
      y tick label style={
        /pgf/number format/.cd,
            fixed,
            fixed zerofill,
            precision=1,
        /tikz/.cd
        },
    ]
        \addplot [
            ultra thick, color = cRed, mark=*,
        ]  coordinates {(Direct,4.782)(\textsc{PMI}-1,4.669) (\textsc{PMI}-3,4.559) (\textsc{PMI}-6,4.507)};
        \addplot [
            only marks, ultra thick, color = cRed, mark=o, mark size=8pt,
        ]  coordinates {(\textsc{PMI}-1,4.669) (\textsc{PMI}-3,4.559) (\textsc{PMI}-6,4.507)};
    \end{axis}
    \draw [thick, cRed, dashed] (0,5.2)--(5.57,5.2);

    \begin{axis}[
      x=4em,
      at={(22em,0)},
      ybar,
      bar width=.58cm,
      bar shift=0pt,
      axis y line*=left,
      ymin=76,
      ymax=84,
      xtick={1,2,3,4,5},
      x tick style={draw=none},
      xticklabels={Direct,\textsc{PMI}-1,\textsc{PMI}-3,\textsc{PMI}-6},
      xticklabel style={rotate=20, anchor=center, xshift=-0.5em, yshift=-1.5em},
      xlabel=(b) $De \rightarrow Fr$$^\ast$,
      x label style={yshift=-1.0em},
      y tick label style={
        /pgf/number format/.cd,
            fixed,
            fixed zerofill,
            precision=1,
        /tikz/.cd
        },
    ]
        \addplot [
            draw = none,
            fill = cGray
        ]  coordinates {(1,77.1)};
        \addplot [
            draw = none,
            fill = cGreen
        ]  coordinates {(2,80.3) (3,82.6) (4,82.5)};
    \end{axis}
    
    \begin{axis}[
      x=4em,
      at={(22em,0)},
      axis y line*=right,
      axis x line=none,
      xtick=data,
      xticklabel style={rotate=20, anchor=center, xshift=0em, yshift=-1.0em},
      xlabel=$x\_label$,
      symbolic x coords={Direct, \textsc{PMI}-1, \textsc{PMI}-3, \textsc{PMI}-6},
      y tick label style={
        /pgf/number format/.cd,
            fixed,
            fixed zerofill,
            precision=1,
        /tikz/.cd
        },
    ]
        \addplot [
            ultra thick, color = cRed, mark=*,
        ]  coordinates {(Direct,4.905) (\textsc{PMI}-1,4.736) (\textsc{PMI}-3,4.678) (\textsc{PMI}-6,4.617)};
        \addplot [
            only marks, ultra thick, color = cRed, mark=o, mark size=8pt,
        ]  coordinates {(\textsc{PMI}-1,4.736) (\textsc{PMI}-3,4.678) (\textsc{PMI}-6,4.617)};
    \end{axis}
    \draw [thick, cRed, dashed] (22em,5.2)--(36.5em,5.2);

    \begin{axis}[
      x=4em,
      at={(44em,0)},
      ybar,
      bar width=.58cm,
      bar shift=0pt,
      axis y line*=left,
      ymin=84.2,
      ymax=85.0,
      xtick={1,2,3,4,5},
      x tick style={draw=none},
      xticklabels={Direct,\textsc{PMI}-1,\textsc{PMI}-3,\textsc{PMI}-6},
      xticklabel style={rotate=20, anchor=center, xshift=-0.5em, yshift=-1.5em},
      xlabel=(b) $De \rightarrow En$$^\dag$,
      x label style={yshift=-1.0em},
      y tick label style={
        /pgf/number format/.cd,
            fixed,
            fixed zerofill,
            precision=1,
        /tikz/.cd
        },
    ]
        \addplot [
            draw = none,
            fill = cGray
        ]  coordinates {(1,84.4)};
        \addplot [
            draw = none,
            fill = cGreen
        ]  coordinates {(2,84.8) (3,84.9) (4,84.9)};
    \end{axis}
    
    \begin{axis}[
      x=4em,
      at={(44em,0)},
      axis y line*=right,
      axis x line=none,
      xtick=data,
      xticklabel style={rotate=20, anchor=center, xshift=0em, yshift=-1.0em},
      xlabel=$x\_label$,
      ytick={27.6,27.7,27.8},
      symbolic x coords={Direct, \textsc{PMI}-1, \textsc{PMI}-3, \textsc{PMI}-6},
      y tick label style={
        /pgf/number format/.cd,
            fixed,
            fixed zerofill,
            precision=1,
        /tikz/.cd
        },
    ]
        \addplot [
            ultra thick, color = cRed, mark=*,
        ]  coordinates {(Direct,27.776) (\textsc{PMI}-1,27.807) (\textsc{PMI}-3,27.769) (\textsc{PMI}-6,27.586)};
        \addplot [
            only marks, ultra thick, color = cRed, mark=triangle, mark size=10pt,
        ]  coordinates {(\textsc{PMI}-1,27.807)};
        \addplot [
            only marks, ultra thick, color = cRed, mark=o, mark size=8pt,
        ]  coordinates {(\textsc{PMI}-3,27.769) (\textsc{PMI}-6,27.586)};
    \end{axis}
    \draw [thick, cRed, dashed] (44em,4.53)--(58.5em,4.53);

    \begin{axis}[
      x=4em,
      at={(66em,0)},
      ybar,
      bar width=.58cm,
      bar shift=0pt,
      axis y line*=left,
      ymin=81.5,
      ymax=83.5,
      xtick={1,2,3,4,5},
      x tick style={draw=none},
      xticklabels={Direct,\textsc{PMI}-1,\textsc{PMI}-3,\textsc{PMI}-6},
      xticklabel style={rotate=20, anchor=center, xshift=-0.5em, yshift=-1.5em},
      xlabel=(b) $De \rightarrow Fr$$^\dag$,
      x label style={yshift=-1.0em},
      y tick label style={
        /pgf/number format/.cd,
            fixed,
            fixed zerofill,
            precision=1,
        /tikz/.cd
        },
    ]
        \addplot [
            draw = none,
            fill = cGray
        ]  coordinates {(1,81.9)};
        \addplot [
            draw = none,
            fill = cGreen
        ]  coordinates {(2,83.0) (3,83.3) (4,83.3)};
    \end{axis}
    
    \begin{axis}[
      x=4em,
      at={(66em,0)},
      axis y line*=right,
      axis x line=none,
      xtick=data,
      xticklabel style={rotate=20, anchor=center, xshift=0em, yshift=-1.0em},
      xlabel=$x\_label$,
      ylabel=\ref{pgfplots:red line} Activation Proportion (\%),
      ytick={27.2,27.3,27.4},
      symbolic x coords={Direct, \textsc{PMI}-1, \textsc{PMI}-3, \textsc{PMI}-6},
      y tick label style={
        /pgf/number format/.cd,
            fixed,
            fixed zerofill,
            precision=1,
        /tikz/.cd
        },
    ]
        \addplot [
            ultra thick, color = cRed, mark=*,
        ]  coordinates {(Direct,27.332) (\textsc{PMI}-1,27.320) (\textsc{PMI}-3,27.250) (\textsc{PMI}-6,27.189)};
        \addplot [
            only marks, ultra thick, color = cRed, mark=o, mark size=8pt,
        ]  coordinates {(\textsc{PMI}-1,27.320) (\textsc{PMI}-3,27.250) (\textsc{PMI}-6,27.189)};
    \end{axis}
    \draw [thick, cRed, dashed] (66em,5.2)--(80.5em,5.2);

\end{tikzpicture}
\vspace{-0.5em}
\caption{The translation performance and the activation proportion of different prompts on WMT dataset. $\ast$ and $\dag$ stand for Bloomz-176B and Qwen-14B, respectively.}
\label{app fig: The translation performance and the activation proportion of different prompts on WMT dataset}
\end{figure*}

In Figure \ref{app fig: The translation performance and the activation proportion of different prompts on WMT dataset}, we also make statistics of activated neurons in Bloomz-176B and Qwen-14B during the inference on the WMT dataset.

\begin{table}[t!]
\Large
    \centering

 \resizebox{0.8\linewidth}{!}{
    \begin{tabular}{ll|cc|cc}
        \toprule
        \multicolumn{2}{c|}{\textbf{Method}} & \textbf{COMET} & \textbf{AP} & \textbf{COMET} & \textbf{AP} \\ 
        \midrule
        \multicolumn{2}{c|}{\textbf{\textit{Direction}}} & \multicolumn{2}{c|}{\textit{De $\rightarrow$ En}}  & \multicolumn{2}{c}{\textit{De $\rightarrow$ Fr}} \\ 
        
        \midrule
        \multirow{2}{*}{\textbf{w/o FT}} & 0-shot & 89.0 & 28.7 & 84.8 & 27.7 \\ 
         & 5-shot & 89.3 & 28.5 & 85.0 & 27.6 \\ 

        \midrule
        \multirow{2}{*}{\textbf{w/ FT}} & 0-shot & 89.5 & 28.1 & 85.3 & 27.2 \\ 
         & 5-shot & 89.3 & 27.8 & 84.9 & 27.1 \\ 

        \bottomrule
    \end{tabular} 
    }

    \caption{The translation performance and activation proportion (AP) of zero-shot and few-shot on Qwen-14B w/ or w/o fine-tuning (FT).}

    \label{tab:t3}
    
\vspace{-1.1em}

\end{table}
Table \ref{tab:t3} shows the results of few-shot learning, which suggests that it also inhibits neurons, and more neurons are inhibited after the LLM is fine-tuned.

\section{More Analyses}

\subsection{Preliminary Experiments of Constructing PMI}
\label{app:sec1}

\begin{figure}
\centering
\begin{tikzpicture} [scale=0.42]
    \tikzstyle{every node}=[font=\Large]
    
    \node [rectangle,draw=black,inner sep=2pt,minimum height=1.1em,minimum width=17.15em,font=\small,anchor=north,align=center] (final) at (20.4em,20em){};
    
    \draw [anchor=center, line width=2pt, Summer] (1.3,7.15)--(3.3,7.15);
    \node [anchor=center] at (4.8,7.2){\scriptsize PMI};
    \draw [anchor=center, line width=2pt, Mizuiro] (9.5,7.15)--(11.5,7.15);
    \node [anchor=center] at (13.3,7.2){\scriptsize Pivot};

    \node (titlea)[anchor=center,font=\scriptsize] at(3.5,-2){(a) BLEU};
    \begin{axis} [
        at={(0,0)},
        xtick={1,2,3,4},
        x tick style={draw=none},
        xticklabels={Fr, Zh, Uk, It},
        nodes near coords,
        nodes near coords style={
            font=\large,
            /pgf/number format/.cd,
            fixed zerofill,
            precision=1,
        },
        xmajorgrids,
        ymajorgrids,
        grid style={draw=black!10, dashed},
    ]
    
    \addplot [
        line width=3pt,
        mark=o,
        draw = Summer,
    ]  coordinates {(1,51.7)(2,47.1)(3,49.1)(4,46.9)};
    \label{pgfplots:PMI line}
    \addplot [
        line width=3pt,
        mark=o,
        draw = Mizuiro,
    ]  coordinates {(1,46.9)(2,32.0)(3,39.9)(4,34.8)};
    
    \end{axis} 


    \node (titleb)[anchor=center,font=\scriptsize] at(12,-2){(b) COMET};
    \begin{axis} [
        at={(23em,0)},
        xtick={1,2,3,4},
        x tick style={draw=none},
        xticklabels={Fr, Zh, Uk, It},
        nodes near coords,
        nodes near coords style={
            font=\normalsize,
            /pgf/number format/.cd,
            fixed zerofill,
            precision=1,
        },
        xmajorgrids,
        ymajorgrids,
        grid style={draw=black!10, dashed},
    ]
    
    \addplot [
        line width=3pt,
        mark=o,
        draw = Summer,
    ]  coordinates {(1,90.4)(2,89.9)(3,90.1)(4,90.0)};
    \addplot [
        line width=3pt,
        mark=o,
        draw = Mizuiro,
    ]  coordinates {(1,89.6)(2,87.4)(3,87.2)(4,88.3)};
    
    \end{axis} 
\end{tikzpicture}
\caption{Examining the factor of selecting parallel languages for PMI. The experiment is conducted on FLORES-200 De $\rightarrow$ En in PMI-1.}
\label{app fig: language_selection}
\end{figure}
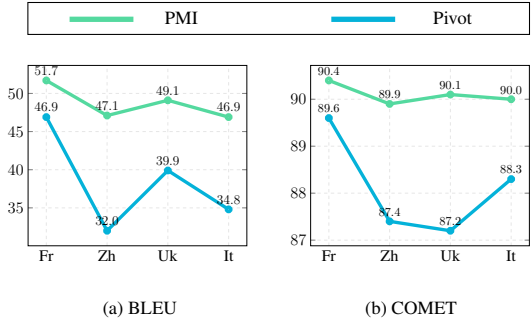

\paragraph{Choose the parallel language that LLMs can understand.} We test the impact of selecting parallel languages on the PMI-1 translating De $\rightarrow$ En of the FLORES-200, where Zh, Fr, Uk, and It are selected as the parallel languages. By comparing the results of translating them to English, we examine the model's understanding of these languages. In Figure \ref{app fig: language_selection}, experimental results show that PMI-1 achieves better performance when the score of pivot translation is high and returns worse results when the score of pivot translation is low. This suggests that choosing parallel languages that the model comprehends better can bring more benefits for PMI.

\begin{figure}
\centering
\begin{tikzpicture} [scale=0.4]
\tikzstyle{every node}=[font=\large]

\begin{axis} [
    at={(0,0)},
    ybar,
    axis y line*=left,
    ylabel=\ref{pgfplots:blue bar} \ref{pgfplots:green bar}  COMET,
    ymin=88.5,
    ymax=90,
    xtick = data,
    xticklabel style={rotate=15, anchor=center, xshift=0em, yshift=-1.0em},
    bar width=.58cm,
    enlarge x limits=0.1,
    xmajorgrids,
    ymajorgrids,
    grid style={draw=black!10, dashed},
    y tick label style={
        /pgf/number format/.cd,
            fixed,
            fixed zerofill,
            precision=1,
        /tikz/.cd
    },
    nodes near coords,
    symbolic x coords={NLLB-1.3B, Qwen-14B, NLLB-54B, ChatGPT, GPT-4}
    ]
    
    \addplot [
        draw = cBlue,
        fill = cBlue
    ]  coordinates {(NLLB-1.3B,89.1) (Qwen-14B,89.1) (NLLB-54B,89.5) (ChatGPT,89.6) (GPT-4,89.6)};
    \label{pgfplots:blue bar}
    
\end{axis} 

\begin{axis}[
    at={(0,0)},
    axis y line*=right,
    axis x line=none,
    xtick=data,
    symbolic x coords={NLLB-1.3B, Qwen-14B, NLLB-54B, ChatGPT, GPT-4},
    y tick label style={
        /pgf/number format/.cd,
            fixed,
            fixed zerofill,
            precision=1,
        /tikz/.cd
    },
    ]
    \addplot [
        ultra thick, color = cRed, mark=*,
    ]   coordinates {(NLLB-1.3B,85.0) (Qwen-14B,85.4) (NLLB-54B,86.7) (ChatGPT,86.9) (GPT-4,87.1)};
\end{axis}

\node (titlea)[anchor=center,font=\scriptsize] at(3.5,-2){(a) Qwen-14B};

\begin{axis} [
    at={(23em,0)},
    ybar,
    axis y line*=left,
    ymin=89.2,
    ymax=90,
    xtick = data,
    xticklabel style={rotate=15, anchor=center, xshift=0em, yshift=-1.0em},
    bar width=.58cm,
    enlarge x limits=0.1,
    xmajorgrids,
    ymajorgrids,
    grid style={draw=black!10, dashed},
    y tick label style={
        /pgf/number format/.cd,
            fixed,
            fixed zerofill,
            precision=1,
        /tikz/.cd
    },
    nodes near coords,
    symbolic x coords={NLLB-1.3B, Qwen-14B, NLLB-54B, ChatGPT, GPT-4},
    ]
    
    \addplot [
        draw = cGreen,
        fill = cGreen
    ] coordinates {(NLLB-1.3B,89.6) (Qwen-14B,89.5) (NLLB-54B,89.7) (ChatGPT,89.5) (GPT-4,89.6)};

\end{axis} 

\begin{axis}[
    at={(23em,0)},
    axis y line*=right,
    axis x line=none,
    xtick=data,
    ylabel=\ref{pgfplots:red line} Average COMET,
    symbolic x coords={NLLB-1.3B, Qwen-14B, NLLB-54B, ChatGPT, GPT-4},
    y tick label style={
        /pgf/number format/.cd,
            fixed,
            fixed zerofill,
            precision=1,
        /tikz/.cd
    },
    ]
    \addplot [
        ultra thick, color = cRed, mark=*,
    ]   coordinates {(NLLB-1.3B,85.0) (Qwen-14B,85.4) (NLLB-54B,86.7) (ChatGPT,86.9) (GPT-4,87.1)};
\end{axis}

\node (titleb)[anchor=center,font=\scriptsize] at(12,-2){(b) ChatGPT};

\end{tikzpicture}
\vspace{-1.3em}
\caption{Examining the factor of translation quality for PMI. This experiment is conducted on FLORES-200 De $\rightarrow$ En in PMI-3. Each point on the red line represents the average COMET score of translating German to the three parallel languages by a translation system, reflecting the different translation qualities of parallel languages.}
\label{app fig: the quality of the translations and their performance}
\end{figure}
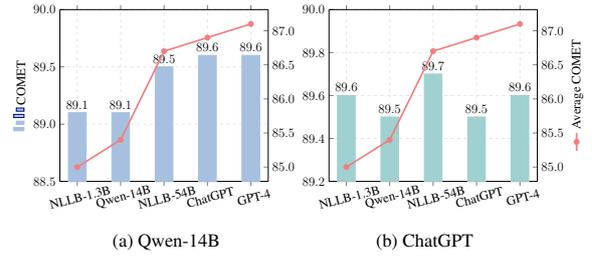

\paragraph{Provide the highest quality translations as far as you can.} Here, we utilize some translation systems with different performances to construct the parallel input of PMI in various qualities, including NLLB-1.3B, NLLB-54B, Qwen-14B, ChatGPT, and GPT-4. Experiments are conducted on both Qwen-14B and ChatGPT. In Figure \ref{app fig: the quality of the translations and their performance}, translation systems are arranged in the ascending order of their translation performance according to the curve, and the results show that higher quality of translations can result in larger improvements.

\begin{table}[t!]
\small
    \centering

 \resizebox{0.7\linewidth}{!}{
    \begin{tabular}{llc}
        \toprule
        \textbf{Method} & \textbf{Input} & \textbf{COMET} \\

        \midrule
        \multirow{4}{*}{\textbf{Direct}} & De & 89.5 \\
        & Es & 87.4 \\
        & Ru & 86.9 \\
        & Zh & 86.9 \\
        
        \midrule
        \multicolumn{3}{c}{\textit{German $\rightarrow$ English}} \\
        \midrule
        \multirow{3}{*}{\textbf{\textsc{PMI}-3}} & \textbf{De + Zh + Ru + Es} & \textbf{90.5} \\
        & De + Zh + Es + Ru & 90.4 \\
        & De + Ru + Es + Zh & 90.3 \\

        \midrule
        \multicolumn{3}{c}{\textit{Chinese $\rightarrow$ English}} \\
        \midrule
        \multirow{3}{*}{\textbf{\textsc{PMI}-3}} & \textbf{Zh + Ru + De + Es} & \textbf{90.3} \\
        & Zh + Ru + Es + De & 90.2 \\
        & Zh + Es + De + Ru & 90.0 \\

        \bottomrule
    \end{tabular}
    }

    \caption{Examining the factor of language order for \textsc{PMI}. The experiment is conducted on FLORES-200 and Qwen-14B.}

    \label{app tab: Examining the impact of language order}

\end{table}

\paragraph{Place better understood language at the head and tail of the input sequence.} We test the performance of PMI prompts with identical parallel texts but in different language order, and conduct experiments on De $\rightarrow$ En and Zh $\rightarrow$ En of the FLORES-200 using Qwen-14B. Results in Table \ref{app tab: Examining the impact of language order} show that an LLM yields superior results when German is placed at the beginning and Spanish is placed at the end. Considering German and Spanish achieve higher scores than other languages, we can infer that it is better to place the language better understood by the model at both ends of the input sequence.

\begin{table*}[t!]
\LARGE
    \centering

 \resizebox{1.0\linewidth}{!}{
    \begin{tabular}{ll|cc|cc|cc|cc|cc|cc}
        \toprule
        \multicolumn{2}{c|}{\textbf{System}} & \textbf{BLEU} & \textbf{COMET} & \textbf{BLEU} & \textbf{COMET} & \textbf{BLEU} & \textbf{COMET} & \textbf{BLEU} & \textbf{COMET} & \textbf{BLEU} & \textbf{COMET} & \textbf{BLEU} & \textbf{COMET}  \\
        \midrule
        \multicolumn{2}{c|}{\textbf{\textit{Direction}}} & \multicolumn{2}{c|}{\textit{De $\rightarrow$ En}} & \multicolumn{2}{c|}{\textit{Zh $\rightarrow$ En}}  & \multicolumn{2}{c|}{\textit{De $\rightarrow$ Fr}}  & \multicolumn{2}{c|}{\textit{En $\rightarrow$ De}} & \multicolumn{2}{c|}{\textit{En $\rightarrow$ Zh}} & \multicolumn{2}{c}{\textit{Is $\rightarrow$ En}} \\
        \midrule
        \multicolumn{2}{c|}{\textbf{\textit{Parallel Languages}}} & \multicolumn{2}{c|}{\textit{Es Ru Fr Zh Ja Cs}} & \multicolumn{2}{c|}{\textit{Es Ru Fr Ja Cs De}}  & \multicolumn{2}{c|}{\textit{En Ru Es Zh It Cs}}  & \multicolumn{2}{c|}{\textit{Es Ru Fr Zh Ja Cs}} & \multicolumn{2}{c|}{\textit{Es Ru Fr Ja Cs De}} & \multicolumn{2}{c}{\textit{Es Ru Fr it Cs De}} \\

        \midrule
        \multirow{3}{*}{\textbf{ChatGPT}} & Direct (1-shot) $^\ast$ & 29.8 & 82.7 & 24.7 & 81.9 & 38.6 & 84.1 & 34.5 & 87.2 & 43.8 & 87.2 & 35.6 & 84.5  \\
         & Direct (5-shot) $^\ast$ & \textbf{32.9} & 85.6 & \textbf{25.4} & 82.6 & 40.5 & 84.5 & 34.7 & 87.4 & 44.4 & \textbf{87.4} & 37.9 & 85.9  \\
         & \cellcolor{shadow}\textsc{PMI} (5-shot) $^\ast$ & \cellcolor{shadow}32.8 & \cellcolor{shadow}\textbf{85.7} & \cellcolor{shadow}24.9 & \cellcolor{shadow}\textbf{82.9} & \cellcolor{shadow}\textbf{41.5} & \cellcolor{shadow}\textbf{84.7} & \cellcolor{shadow}\textbf{34.8} & \cellcolor{shadow}\textbf{87.6} & \cellcolor{shadow}\textbf{45.1} & \cellcolor{shadow}87.3 & \cellcolor{shadow}\textbf{39.3} & \cellcolor{shadow}\textbf{86.7}  \\

        \midrule
        \multirow{3}{*}{\textbf{Qwen-14B}} & Direct (0-shot) $^\dag$ & 30.4 & 84.4 & 23.7 & 80.8 & 34.2 & 81.9 & 29.6 & 85.3 & 45.2 & \textbf{87.6} & 18.4 & 69.7  \\

         & Direct (5-shot) $^\ast$ & 31.5 & 84.7 & 24.0 & 80.8 & 33.0 & 81.8 & 29.3 & 84.9 & \textbf{45.4} & 87.3 & 19.6 & 71.9  \\
        & \cellcolor{shadow}\textsc{PMI} (0-shot) $^\dag$ & \cellcolor{shadow}\textbf{31.6} & \cellcolor{shadow}\textbf{84.9} & \cellcolor{shadow}\textbf{24.3} & \cellcolor{shadow}\textbf{82.0} & \cellcolor{shadow}\textbf{38.4} & \cellcolor{shadow}\textbf{83.4} & \cellcolor{shadow}\textbf{30.0} & \cellcolor{shadow}\textbf{85.8} & \cellcolor{shadow}45.1 & \cellcolor{shadow}\textbf{87.6} & \cellcolor{shadow}\textbf{37.9} & \cellcolor{shadow}\textbf{85.7}  \\

         \midrule
        \multirow{3}{*}{\textbf{ALMA-13B}} & Direct (0-shot) $^\dag$ & 28.1 & 83.8 & 21.6 & 79.6 & 27.1 & 79.2 & 29.6 & 85.5 & 36.9 & 85.8 & 34.0 & 85.8  \\
         & Paper Reported $^\ast$ & 30.7 & 84.4 & \textbf{24.7} & 79.9 & - & - & 31.4 & 85.5 & \textbf{39.1} & 85.8 & 36.5 & 86.3   \\
        & \cellcolor{shadow}\textsc{PMI} (0-shot) $^\dag$ & \cellcolor{shadow}\textbf{30.8} & \cellcolor{shadow}\textbf{85.0} & \cellcolor{shadow}23.8 & \cellcolor{shadow}\textbf{81.8} & \cellcolor{shadow}\textbf{33.3} & \cellcolor{shadow}\textbf{81.5} & \cellcolor{shadow}\textbf{29.9} & \cellcolor{shadow}\textbf{86.0} & \cellcolor{shadow}36.9 & \cellcolor{shadow}\textbf{86.0} & \cellcolor{shadow}\textbf{38.3} & \cellcolor{shadow}\textbf{86.5}  \\

         \midrule
        \multirow{3}{*}{\textbf{Bloomz-176B}} & Direct (0-shot) $^\ast$ & 24.0 & 78.4 & 16.0 & 76.4 & 27.3 & 77.1 & 13.0 & 70.7 & 29.5 & 83.9 & \ \ 5.6 & 53.8  \\
         & Direct (5-shot) $^\ast$ & 23.1 & 79.7 & 14.5 & 77.3 & 25.9 & 77.2 & \textbf{16.1} & \textbf{74.1} & \textbf{33.5} & \textbf{85.2} & \ \ 5.1 & 56.1  \\
        & \cellcolor{shadow}\textsc{PMI} (0-shot) $^\ast$ & \cellcolor{shadow}\textbf{28.2} & \cellcolor{shadow}\textbf{83.9} & \cellcolor{shadow}\textbf{21.7} & \cellcolor{shadow}\textbf{81.4} & \cellcolor{shadow}\textbf{36.6} & \cellcolor{shadow}\textbf{82.9} & \cellcolor{shadow}16.0 & \cellcolor{shadow}73.9 & \cellcolor{shadow}32.4 & \cellcolor{shadow}84.7 & \cellcolor{shadow}\textbf{34.0} & \cellcolor{shadow}\textbf{84.2}  \\

        \bottomrule
    \end{tabular} 
    }

    \caption{Comparing the performance of few-shot and \textsc{PMI}. In fairness, the results of few-shot come from models without fine-tuning, i.e., the official release. $\dag$ and $\ast$ represent whether the prompt is fed to a model that has been fine-tuned or not, respectively.}

    \label{app tab: Comparing the performance between few-shot and PiM}
    

\end{table*}

\subsection{Comparing the Performance Between Few-shot Learning and PMI}
To further evaluate the effectiveness of our PMI, here we compare the results of PMI with those of few-shot learning. Notably, since our fine-tuning data is constructed by zero-shot instructions, which hurts the performance of few-shot learning evaluated on these fine-tuned models \cite{DBLP:conf/emnlp/AlvesGAPRSCM23}, hence we conduct experiments of few-shot learning on original models, i.e., the officially released weights without being fine-tuned. As shown in Table \ref{app tab: Comparing the performance between few-shot and PiM}, PMI also outperforms few-shot learning.

\begin{table}[t!]
\large
    \centering

 \resizebox{0.9\linewidth}{!}{
    \begin{tabular}{lllll}
        \toprule
        \textbf{Method} & \textbf{Time Cost} & \makecell[l]{\textbf{Increase} \\  \textbf{Rate (\%)}} & \textbf{BLEU} & \makecell[l]{\textbf{Increase} \\  \textbf{Rate (\%)}} \\

        \midrule
        Direct & 189.4s & - & 45.2 & - \\
        \textsc{PMI}-1 & 249.7s & 31.8 & 47.9 & 5.9 \\
        \textsc{PMI}-3 & 397.9s & 110.1 & 56.2 & 24.3 \\
        \textsc{PMI}-5 & 507.3s & 167.8 & 56.5 & 25.0 \\

        \bottomrule
    \end{tabular} 
    }

    \caption{The inference speed and performance gain of \textsc{PMI} with different amount of parallel languages.}

    \label{app tab: Sacrificing a certain level of reasoning speed can achieve outstanding performance gains}

\end{table}

\begin{table}[t!]
\Large
    \centering

 \resizebox{0.85\linewidth}{!}{
    \begin{tabular}{l|cc|cc}
        \toprule
        \textbf{System} & \textbf{BLEU} & \textbf{COMET} & \textbf{BLEU} & \textbf{COMET} \\
        
        \midrule
        \textbf{\textit{Direction}} & \multicolumn{2}{c|}{\textit{De $\rightarrow$ En}}  & \multicolumn{2}{c}{\textit{Zh $\rightarrow$ En}}  \\

        \midrule
        \textbf{Direct} & 24.8 & 83.0 & 12.1 & 76.8 \\
        \textbf{Pivot}  & 23.4 & 83.4 & \textbf{17.2} & 80.7 \\
        \cellcolor{shadow}\textbf{PMI} & \cellcolor{shadow}\textbf{25.2} & \cellcolor{shadow}\textbf{84.4} & \cellcolor{shadow}17.0 & \cellcolor{shadow}\textbf{81.1} \\
        
        \midrule
        \textbf{\textit{Direction}} & \multicolumn{2}{c|}{\textit{En $\rightarrow$ De}}  & \multicolumn{2}{c}{\textit{En $\rightarrow$ Zh}}  \\

        \midrule
        \textbf{Direct} & 22.9 & 81.5 & 36.1 & 85.9  \\
        \textbf{Pivot}  & 21.0 & 82.1 & 35.7 & 85.2  \\
        \cellcolor{shadow}\textbf{PMI} & \cellcolor{shadow}\textbf{23.2} & \cellcolor{shadow}\textbf{83.4} & \cellcolor{shadow}\textbf{39.8} & \cellcolor{shadow}\textbf{86.5} \\

        \bottomrule
    \end{tabular} 
    }

    \caption{Experiments of Qwen1.5-14B on the WMT dataset.}

    \label{tab:Experiments of Qwen1.5-14B on the WMT dataset}
    
\vspace{-0.7em}

\end{table}

\begin{table}[t!]
\Large
    \centering

 \resizebox{0.85\linewidth}{!}{
    \begin{tabular}{l|cc|cc}
        \toprule
        \textbf{System} & \textbf{BLEU} & \textbf{COMET} & \textbf{BLEU} & \textbf{COMET} \\
        
        \midrule
        \textbf{\textit{Direction}} & \multicolumn{2}{c|}{\textit{Zh $\rightarrow$ En}}  & \multicolumn{2}{c}{\textit{De $\rightarrow$ Fr}}  \\

        \midrule
        \textbf{Direct} & \textbf{23.7} & 80.8 & 34.2 & 81.9 \\
        \textbf{Pivot}  & 15.9 & 78.7 & 36.2 & 81.3 \\
        \cellcolor{shadow}\textbf{PMI} & \cellcolor{shadow}22.1 & \cellcolor{shadow}\textbf{80.9} & \cellcolor{shadow}37.6 & \cellcolor{shadow}\textbf{82.7} \\
        
        \midrule
        \textbf{\textit{Direction}} & \multicolumn{2}{c|}{\textit{En $\rightarrow$ De}}  & \multicolumn{2}{c}{\textit{En $\rightarrow$ Zh}}  \\

        \midrule
        \textbf{Direct} & \textbf{29.6} & 85.3 & 45.2 & 87.6  \\
        \textbf{Pivot}  & 25.8 & 83.5 & 39.7 & 86.2  \\
        \cellcolor{shadow}\textbf{PMI} & \cellcolor{shadow}\textbf{29.6} & \cellcolor{shadow}\textbf{85.5} & \cellcolor{shadow}\textbf{45.4} & \cellcolor{shadow}\textbf{87.7} \\

        \bottomrule
    \end{tabular} 
    }

    \caption{Augmenting Qwen-14B by the translations from Qwen-14B itself on the WMT dataset.}

    \label{tab:Augmenting Qwen-14B by the translations from Qwen-14B itself on the WMT dataset}
    
\vspace{-0.7em}

\end{table}

\subsection{Effectiveness of PMI on more modern LLMs}
\label{app sec: Effectiveness of PMI on more modern LLMs}
As LLMs develop further, we anticipate that more and more LLMs will benefit from PMI in the future. Here, we make experiments on Qwen1.5-14B, a successor of Qwen-14B. The latter is fine-tuned with PMI prompts in our paper, while the former is the original official version. From Table \ref{tab:Experiments of Qwen1.5-14B on the WMT dataset}, we can see that Qwen1.5-14B responds to PMI prompts without prior fine-tuning and exhibits performance enhancements due to PMI.

\subsection{Self-augmentation}
\label{app sec: Self-augmentation}
In Table \ref{tab:Augmenting Qwen-14B by the translations from Qwen-14B itself on the WMT dataset}, we report the experimental results of prompting Qwen-14B with PMI while the parallel sentence pairs are translated by Qwen-14B itself. Although the improvements resulting from PMI are not as large as those reported in Table \ref{tab:t4}, PMI still outperforms baselines, especially at the COMET score. This further demonstrates the applicability of PMI. We attribute the diminished performance gains to the lower quality of translations produced by Qwen-14B compared to those from GPT-4.

\subsection{Inference Speed}
\label{app sec: Inference Speed}
Since the inference speed of LLMs inevitably slows down as the input sequence lengthens, we also focus on the trade-off between performance and inference speed when increasing the number of parallel languages in the PMI. Here, we conduct experiments on the FLORES-200 De $\rightarrow$ En and Qwen-14B model. Table \ref{app tab: Sacrificing a certain level of reasoning speed can achieve outstanding performance gains} indicates that for every additional parallel language integrated into the PMI input, there is an approximate 30\% increase in time cost, along with a 5\% improvement in performance. Notably, when the number of parallel languages reaches three, the improvement can reach up to 24.34\%. Despite the increased inference cost, it is reasonable and acceptable considering the substantial performance gain.

\section{Details of Experiment Setups}
\label{app sec: Details of Experiment Setups}

\subsection{Downstream tasks}
We introduce the details of the downstream tasks we used here:
\paragraph{Machine Translation} In this task, a source language text is input into the model, which then translates it into a target language.

\paragraph{Nature Language Inference} This task involves inputting a pair of sentences into the model, which then determines and outputs their relational status, such as contradiction, entailment, or neutrality.

\paragraph{Reading Comprehension} This task gives a passage and a question to the model, and then the model answers the question with a `Yes’ or `No’ based on its comprehension.

\paragraph{Text Simplification} This task is to input a complex sentence into the model, and then the model generates a simplified version of the sentence without losing important information or altering its original intent.

\paragraph{Abstractive Summarization} In this task, a long text is input into the model, which then produces a summary in one or two sentences that captures the essence and most critical information of the text.

\subsection{Multilingual LLMs}
\label{app sec: Multilingual LLMs}
Here, we introduce the multilingual LLMs used in our main experiment.

\begin{table}[t!]
\Large
    \centering

 \resizebox{1.0\linewidth}{!}{
    \begin{tabular}{ll|ll|ll}
        \toprule

        \multicolumn{2}{c|}{\textbf{System}} & \textbf{BLEU} & \textbf{COMET} & \textbf{BLEU} & \textbf{COMET} \\
        \midrule
        \multicolumn{2}{c|}{\textbf{\textit{Direction}}} & \multicolumn{2}{c|}{\textit{Fr $\rightarrow$ De}} & \multicolumn{2}{c}{\textit{Fr $\rightarrow$ Es}}  \\

        \midrule
        \multirow{5}{*}{\textbf{ChatGPT}} & Direct & \textbf{30.4}  & \textbf{86.5} & 25.3  & \textbf{86.3}   \\
         & $\textsc{PMI}_{PA}$ & 26.0$^{\downarrow 4.4}$  & 85.7$^{\downarrow 0.8}$ & 24.7$^{\downarrow 0.6}$  & 86.0$^{\downarrow 0.3}$   \\
         & $\textsc{PMI}_{MS}$ & 30.0$^{\downarrow 0.4}$  & 85.6$^{\downarrow 0.9}$ & \textbf{26.1}$^{\uparrow 0.8}$  & 86.2$^{\downarrow 0.1}$  \\
         & $\textsc{PMI}_{ML}$ & \textbf{30.4}$^{\uparrow 0.0}$  & 86.3$^{\downarrow 0.2}$ & 25.5$^{\uparrow 0.2}$  & \textbf{86.3}$^{\uparrow 0.0}$  \\
         & $\textsc{PMI}_{GT}$ & \textit{32.4}  & \textit{86.9} & \textit{27.0}  & \textit{86.8}  \\

        \midrule
        \multirow{5}{*}{\textbf{Qwen-14b}} & Direct & 25.9  & 84.8 & 24.0  & 85.6   \\
         & $\textsc{PMI}_{PA}$ & \textbf{28.1}$^{\uparrow 2.2}$  & \textbf{86.0}$^{\uparrow 1.2}$ & 23.5$^{\downarrow 0.5}$  & 85.5$^{\downarrow 0.1}$   \\
         & $\textsc{PMI}_{MS}$ & 27.6$^{\uparrow 1.7}$  & 85.5$^{\uparrow 0.7}$ & \textbf{25.4}$^{\uparrow 1.4}$  & \textbf{86.0}$^{\uparrow 0.4}$  \\
         & $\textsc{PMI}_{ML}$ & 26.8$^{\uparrow 0.9}$  & 85.0$^{\uparrow 0.2}$ & 24.1$^{\uparrow 0.1}$  & 85.8$^{\uparrow 0.2}$   \\
         & $\textsc{PMI}_{GT}$ & \textit{29.6}  & \textit{86.0} & \textit{27.3}  & \textit{86.4}  \\

        \midrule
        \multirow{4}{*}{\textbf{GPT-4}} & Direct & 30.4  & 86.5 & 25.6  & 86.4   \\
         & $\textsc{PMI}_{MS}$ & \textbf{32.1}$^{\uparrow 1.7}$  & \textbf{87.1}$^{\uparrow 0.5}$ & \textbf{26.3}$^{\uparrow 0.7}$  & \textbf{87.0}$^{\uparrow 0.6}$  \\
         & $\textsc{PMI}_{ML}$ & \textbf{32.1}$^{\uparrow 1.7}$  & 86.7$^{\uparrow 0.2}$ & 25.9$^{\uparrow 0.3}$  & 86.5$^{\uparrow 0.1}$   \\
         & $\textsc{PMI}_{GT}$ & \textit{35.8}  & \textit{87.7} & \textit{28.4}  & \textit{87.3}  \\

        \bottomrule
    \end{tabular} 
    }

    \caption{Supplement results of the ablation study.}

    \label{app tab: Supplement results of the ablation study}

\end{table}

\begin{table*}[t!]
\small
    \centering

 \resizebox{0.8\linewidth}{!}{
    \begin{tabular}{llcccccc}
        \toprule
        \multirow{2.5}{*}{\textbf{Model}} & \multirow{2.5}{*}{\textbf{Task}} & \multicolumn{3}{c}{\textbf{Training Super Parameters}} & \multicolumn{3}{c}{\textbf{Training Data}} \\
        \cmidrule(lr){3-5} \cmidrule(lr){6-7}
        &  & \textbf{Batch Size} & \textbf{Epoch} & \textbf{Learning Rate} & \textbf{Ratio} & \textbf{Size}\\

        \midrule
        \multirow{5}{*}{\textbf{Qwen-7B}} & Machine Translation & 16 & 1 & 2e-5 & 1:9 & 4985\\
        & Nature Language Inference & 16 & 2 & 5e-5 & 1:7 & 2000\\
        & Reading Comprehension & 16 & 8 & 8e-5 & 1:5 & 2000\\
        & Text Simplification & 16 & 7 & 7e-5 & 1:5 & 2000\\
        & Abstractive Summarization & 16 & 4 & 1e-5 & 1:9 & 1200\\
        
        \midrule
        \multirow{5}{*}{\textbf{Qwen-14B}} & Machine Translation & 16 & 1 & 2e-5 & 1:9 & 4985\\
        & Nature Language Inference & 16 & 1 & 5e-5 & 1:7 & 2000\\
        & Reading Comprehension & 16 & 9 & 8e-5 & 1:7 & 2000\\
        & Text Simplification & 16 & 7 & 7e-5 & 1:5 & 2000\\
        & Abstractive Summarization & 16 & 4 & 7e-5 & 1:7 & 1200\\

        \midrule
        \multirow{5}{*}{\textbf{ALMA-13B}} & Machine Translation & 16 & 1 & 5e-5 & 1:9 & 4985\\
        & Nature Language Inference & 16 & 6 & 5e-5 & 1:7 & 2000\\
        & Reading Comprehension & 16 & 6 & 8e-5 & 1:7 & 2000\\
        & Text Simplification & 16 & 8 & 7e-5 & 1:9 & 2000\\
        & Abstractive Summarization & 16 & 3 & 2e-4 & 1:9 & 1200\\

        \midrule
        \multirow{4}{*}{\textbf{Yi-34B}} & Nature Language Inference & 16 & 3 & 1e-5 & 1:7 & 2000\\
        & Reading Comprehension & 16 & 7 & 8e-5 & 1:9 & 2000\\
        & Text Simplification & 16 & 7 & 5e-5 & 1:9 & 2000\\
        & Abstractive Summarization & 16 & 5 & 7e-5 & 1:9 & 1200\\

        \midrule
        \multirow{2}{*}{\textbf{Qwen-72B}} & Nature Language Inference & 16 & 8 & 1e-5 & 1:7 & 2000\\
        & Reading Comprehension & 16 & 5 & 6e-5 & 1:7 & 2000\\

        \bottomrule
    \end{tabular} 
    }

    \caption{Our training setups. Each model is trained to ensure optimal performance for both the baseline and \textsc{PMI}.}

    \label{app tab: The training setup used in LoRA fine-tuning}

\end{table*}

\begin{table}[htbp]
\LARGE
  \centering
  
 \resizebox{1.0\linewidth}{!}{
 
    \begin{tabular}{l|cc|cc|cc}
    \toprule
    \multirow{3.5}{*}{\textbf{Model}} & \multicolumn{2}{c|}{\textbf{WikiAuto}}  & \multicolumn{4}{c}{\textbf{XLSum}}  \\
    \cmidrule(lr){2-3} \cmidrule(lr){4-7}    
    & \multicolumn{2}{c|}{\textit{En}} & \multicolumn{2}{c|}{\textit{Es}} & \multicolumn{2}{c}{\textit{Ru}} \\
    \cmidrule(lr){2-3} \cmidrule(lr){4-7}  
    & Pivot & SARI & Pivot & R2/RL & Pivot & R2/RL  \\

    \midrule    
    \multirow{3}{*}{\textbf{Qwen-7B}} & \textbf{Fr} & \textbf{43.2} & \textbf{Fr} & \textbf{9.4/22.7} & \textbf{Es} & \textbf{41.1/38.5} \\
    & De & 43.1 & - & - & - & - \\
    & Es & 43.0 & - & - & - & - \\

    \midrule    
    \multirow{3}{*}{\textbf{Qwen-14B}} & Fr & 43.6 & \textbf{Fr} & \textbf{9.0/21.4} & \textbf{Es} & \textbf{40.2/38.3} \\
    & De & 43.1 & - & - & - & - \\
    & \textbf{Es} & \textbf{43.8} & - & - & - & - \\

    \midrule    
    \multirow{3}{*}{\textbf{ALMA-13B}} & Fr & 43.1 & \textbf{Fr} & \textbf{10.4/23.0} & \textbf{Es} & \textbf{44.3/41.2} \\
    & \textbf{De} & \textbf{43.2} & - & - & - & - \\
    & \textbf{Es} & \textbf{43.2} & - & - & - & - \\

    \midrule    
    \multirow{3}{*}{\textbf{Yi-34B}} & \textbf{Fr} & \textbf{43.5} & \textbf{Fr} & \textbf{10.6/23.3} & \textbf{Es} & \textbf{41.7/38.8} \\
    & De & 43.3 & - & - & - & - \\
    & Es & 42.4 & - & - & - & - \\
    
    \bottomrule
    \end{tabular}    
}

  \caption{Full experimental results of pivot prompts on WikiAuto and XLSum dataset. The best results of each group are in \textbf{bold}.}
  \label{app tab: Full experimental results of pivot prompts on WikiAuto and XLSum dataset}
\end{table}

\paragraph{ChatGPT:} the most capable GPT-3.5 model, which performs impressively on rich-resource languages. We use the \texttt{gpt-3.5-turbo-0613} API.

\paragraph{LLaMA3:} an open-source multilingual LLM which is pre-trained with 15 trillion tokens and demonstrated superior performance across multiple benchmarks \cite{llama3modelcard}.

\paragraph{Bloomz:} a fine-tuned version of Bloom \cite{DBLP:journals/corr/abs-2211-05100}, and we conduct experiments on the largest \texttt{bloomz} containing 176B parameters.

\paragraph{Qwen:} open-source models which are trained up to 3 trillion tokens of multilingual data with competitive performance on various tasks \cite{DBLP:journals/corr/abs-2309-16609}. We do evaluations on three models, including Qwen-7B (\texttt{Qwen-7B-Chat}), Qwen-14B (\texttt{Qwen-14B-Chat}) and Qwen-72B (\texttt{Qwen-72B-Chat}).

\paragraph{ALMA:} a multilingual LLaMA-2 \cite{DBLP:journals/corr/abs-2307-09288} produced by continually pre-training and specially instruction-tuning on the MT task \cite{DBLP:journals/corr/abs-2309-11674}. We conduct experiments on \texttt{ALMA-13B}.

\paragraph{Yi:} an open-source model which is mainly trained on English and Chinese corpus achieving competitive performance on multilingual tasks \cite{Yi}. We assess the effectiveness of PMI on Yi-34B (\texttt{Yi-34B-Chat}).

\paragraph{mT0:} an instruction-tuned version of mT5 \cite{DBLP:conf/naacl/XueCRKASBR21}, we choose the mT0-13B (\texttt{mt0-xxl}) as it supports 46 languages.

\subsection{Training Setups}
Limited by parameters and training data, it might be a challenge for every LLM to understand PMI prompts inherently. To address this, we conducted training data and fine-tuned the models, which seemed confused when facing the PMI prompt. Specifically, we leveraged LLaMA-Factory\footnote{\url{https://github.com/hiyouga/LLaMA-Factory}} \cite{llama-factory} and the LoRA technology to train models, where we set the LoRA-rank to 8, LoRA-alpha to 32, and dropout to 0.1. Since the different number of trainable parameters in the LoRA module, we applied different training strategies to ensure that every model can adequately understand prompts of various tasks. These settings are detailed in Table \ref{app tab: The training setup used in LoRA fine-tuning}.

\subsection{Details of the Fine-tuning Datasets}
We constructed our fine-tuning dataset based on the training or development datasets of these tasks for both conventional and PMI prompts in zero-shot style. There are two factors, including the ratio of baseline to PMI data and the size of the training dataset, which are detailed in Table \ref{app tab: The training setup used in LoRA fine-tuning}.

\subsection{Decoding Setups}
We kept consistent super parameters during the inference stage of every LLM, i.e., we set the decoding batch size to 4 and the temperature to 0.01 in order to ensure the reproducibility of the results.

\section{Full Experimental Results of Pivot Prompts}
\label{app sec: Full Experimental Results of Pivot Prompts}
We have reported the results of pivot prompts with the highest score in the table of the main experiment. In Tables \ref{app tab: Full experimental results of pivot prompts on WikiAuto and XLSum dataset}, \ref{app tab: Full experimental results of pivot prompts on WMT dataset}, and \ref{app tab: Full experimental results of pivot prompts on RTE, XNLI and BoolQ dataset}, we list all the results of the pivot prompts.

\begin{table*}[htbp]
\Large
  \centering
  
 \resizebox{1.0\linewidth}{!}{
 
    \begin{tabular}{c|ccc|ccc|ccc|ccc|ccc|ccc}
    \toprule
    {\textbf{Model}} & \textbf{Pivot} & \textbf{BLEU}  & \textbf{COMET} & \textbf{Pivot} & \textbf{BLEU}  & \textbf{COMET} & \textbf{Pivot} & \textbf{BLEU}  & \textbf{COMET} & \textbf{Pivot} & \textbf{BLEU}  & \textbf{COMET} & \textbf{Pivot} & \textbf{BLEU}  & \textbf{COMET} & \textbf{Pivot} & \textbf{BLEU}  & \textbf{COMET} \\
    
    \midrule        

    \textbf{\textit{Direction}} & \multicolumn{3}{c|}{\textit{De $\rightarrow$ En}} & \multicolumn{3}{c|}{\textit{Zh $\rightarrow$ En}} & \multicolumn{3}{c|}{\textit{De $\rightarrow$ Fr}} & \multicolumn{3}{c|}{\textit{En $\rightarrow$ De}} & \multicolumn{3}{c|}{\textit{En $\rightarrow$ Zh}} & \multicolumn{3}{c}{\textit{Is $\rightarrow$ En}} \\

    \midrule
    
    \multicolumn{1}{c|}{\multirow{6}{*}{\textbf{ChatGPT}}} & \textbf{Es} & \textbf{28.5} & \textbf{84.0} & \textbf{Es} & \textbf{21.6} & \textbf{81.9} & \textbf{En} & \textbf{40.4} & \textbf{84.0} & Es    & 30.0    & 85.6  & \textbf{Es} & \textbf{40.3} & \textbf{86.0} & Es    & 34.6  & 85.4 \\
          & Ru    & 25.2  & 83.6  & Ru    & 18.4  & 80.7  & Ru    & 33.1  & 82.6  & Ru    & 27.4  & 86.2  & Ru    & 35.9  & 85.6  & Ru    & 30.5  & 84.6 \\
          & Fr    & 27.3  & 82.6  & Fr    & 16.3  & 76.9  & Es    & 37.0    & 83.3  & \textbf{Fr} & \textbf{30.0} & \textbf{86.4} & Fr    & 36.9  & 85.1  & Fr    & 31.2  & 84.1 \\
          & Zh    & 19.5  & 82.4  & Ja    & 18.5  & 80.1  & Zh    & 25.0    & 80.9  & Zh    & 21.7  & 85.0    & Ja    & 33.4  & 85.0    & It    & 33.0    & 85.0 \\
          & Ja    & 19.5  & 81.7  & Cs    & 18.6  & 80.2  & It    & 37.3  & 83.3  & Ja    & 20.4  & 84.8  & Cs    & 37.2  & 85.4  & Cs    & 27.7  & 81.9 \\
          & Cs    & 25.6  & 81.8  & De    & 20.1  & 81.0    & Cs    & 34.8  & 82.5  & Cs    & 29.0    & 86.1  & De    & 37.9  & 85.9  & \textbf{De} & \textbf{35.0} & \textbf{85.6} \\

    \midrule
    
    \multirow{6}{*}{\textbf{LLaMA3-8B}} & Es    & 26.4  & 83.3  & \textbf{Es} & \textbf{21.3} & \textbf{81.4} & \textbf{En} & \textbf{31.7} & \textbf{80.8} & Es    & 22.8  & 81.8  & \textbf{Es} & \textbf{30.2} & 79.9  & \textbf{Es} & \textbf{32.5} & \textbf{84.9} \\
          & Ru    & 23.3  & 82.7  & Ru    & 17.8  & 79.9  & Ru    & 24.3  & 79.6  & Ru    & 19.6  & 82.1  & Ru    & 26.4  & 81.0  & Ru    & 27.6  & 83.5 \\
          & \textbf{Fr} & \textbf{27.4} & \textbf{83.4} & Fr    & 20    & 80.9  & Es    & 30.7  & 80.5  & \textbf{Fr} & \textbf{24} & \textbf{83.3} & Fr    & 28.8  & 81.0  & Fr    & 32.2  & 85.0 \\
          & Zh    & 18.1  & 81.2  & Ja    & 17.1  & 79.2  & Zh    & 18.1  & 77.3  & Zh    & 14.2  & 80.7  & Ja    & 25.2  & 80.4  & It    & 31    & 84.6 \\
          & Ja    & 16.6  & 80.2  & Cs    & 18.2  & 79.7  & It    & 31.5  & 80.7  & Ja    & 13.5  & 80.5  & Cs    & 28.2  & 81.1  & Cs    & 27.9  & 83.4 \\
          & Cs    & 25.5  & 82.4  & De    & 19.8  & 80.7  & Cs    & 27.5  & 78.8  & Cs    & 21.7  & 82.5  & De    & 29.3  & \textbf{81.7} & De    & 32.4  & 84.8 \\

    \midrule
    
    \multirow{6}{*}{\textbf{Qwen-14B}} & Es    & 28.1  & 83.8  & \textbf{Es} & \textbf{22.4} & \textbf{81.8} & \textbf{En} & \textbf{37.4} & \textbf{82.7} & Es    & 26.5  & 83.7  & \textbf{Es} & \textbf{41.2} & \textbf{86.3} & Es    & 33.7  & 85.2 \\
          & Ru    & 25.0    & 82.9  & Ru    & 19.8  & 80.6  & Ru    & 29.8  & 81.2  & Ru    & 23.5  & 84.1  & Ru    & 38.7  & 86.3  & Ru    & 30.3  & 84.1 \\
          & \textbf{Fr} & \textbf{28.2} & \textbf{84.0} & Fr    & 21.5  & 81.5  & Es    & 34.5  & 82.1  & \textbf{Fr} & \textbf{26.9} & \textbf{84.7} & Fr    & 40.4  & 86.6  & \textbf{Fr} & \textbf{34.1} & \textbf{85.4} \\
          & Zh    & 20.5  & 82.1  & Ja    & 19.1  & 79.8  & Zh    & 24.7  & 79.9  & Zh    & 20.5  & 83.2  & Ja    & 35.6  & 85.5  & It    & 33.0  & 85.0 \\
          & Ja    & 19.2  & 81.3  & Cs    & 19.6  & 80.2  & It    & 34.3  & 82.1  & Ja    & 17.5  & 82.5  & Cs    & 38.5  & 85.5  & Cs    & 29.9  & 84.1 \\
          & Cs    & 25.1  & 82.6  & De    & 20.7  & 81.2  & Cs    & 30.5  & 80.3  & Cs    & 24.3  & 83.8  & De    & 39.1  & 86.3  & De    & 33.8  & 85.2 \\

    \midrule
    
    \multirow{6}{*}{\textbf{ALMA-13B}} & Es    & 25.5  & 83.0  & \textbf{Es} & \textbf{21.7} & \textbf{81.2} & \textbf{En} & \textbf{29.9} & \textbf{80.3} & Es    & 26.2  & 83.7  & Es    & 32.3  & 83.9  & \textbf{Es} & \textbf{32.7} & \textbf{85.2} \\
          & Ru    & 22.8  & 82.5  & Ru    & 18.9  & 80.1  & Ru    & 24.8  & 78.8  & Ru    & 24.6  & 84.8  & Ru    & 31.4  & 84.5  & Ru    & 28.1  & 84.1 \\
          & \textbf{Fr} & \textbf{26.0} & \textbf{83.3} & Fr    & 20.9  & 80.9  & Es    & 29.4  & 79.9  & \textbf{Fr} & \textbf{26.4} & \textbf{84.8} & Fr    & 32.3  & 84.5  & Fr    & 31.7  & 85.0 \\
          & Zh    & 18.1  & 81.0  & Ja    & 16.7  & 78.4  & Zh    & 18.0  & 76.6  & Zh    & 18.8  & 82.9  & Ja    & 28.0  & 82.5  & It    & 31.3  & 84.7 \\
          & Ja    & 16.3  & 79.9  & Cs    & 19.0  & 79.8  & It    & 30.2  & 80.0  & Ja    & 15.8  & 81.2  & Cs    & 32.2  & 84.4  & Cs    & 28.5  & 84.0 \\
          & Cs    & 24.0  & 82.6  & De    & 20.2  & 80.9  & Cs    & 25.7  & 78.2  & Cs    & 25.4  & 84.6  & \textbf{De} & \textbf{32.3} & \textbf{84.6} & De    & 31.8  & 85.1 \\

    \midrule
    
    \multirow{6}{*}{\textbf{mT0-13B}} & \textbf{Es} & \textbf{24.5} & \textbf{82.5} & \textbf{Es} & \textbf{19.3} & \textbf{80.7} & En    & 30.9  & 79.8  & Es    & 17.2  & 77.1  & Es    & 23.4  & 81.9  & \textbf{Es} & \textbf{30.8} & \textbf{84.6} \\
          & Ru    & 21.3  & 81.5  & Ru    & 16.0  & 79.1  & Ru    & 25.7  & 78.6  & Ru    & 15.6  & 77.5  & Ru    & 23.1  & 82.3  & Ru    & 25.9  & 82.9 \\
          & Fr    & 24.5  & 82.4  & Fr    & 18.5  & 80.2  & \textbf{Es} & \textbf{30.5} & \textbf{80.1} & Fr    & 16.8  & 77.2  & Fr    & 23.1  & 82.1  & Fr    & 29.3  & 84.0 \\
          & Zh    & 16.6  & 79.8  & Ja    & 12.9  & 76.8  & Zh    & 18.8  & 76.3  & Zh    & 12.2  & 75.8  & Ja    & 22.3  & 81.9  & It    & 29.6  & 84.1 \\
          & Ja    & 15.6  & 79.3  & Cs    & 16.5  & 79.1  & It    & 30.3  & 80.0  & Ja    & 12.1  & 76.4  & Cs    & 22.9  & 81.6  & Cs    & 27.1  & 83.5 \\
          & Cs    & 22.7  & 81.5  & De    & 17.4  & 79.7  & Cs    & 26.6  & 78.2  & \textbf{Cs} & \textbf{17.4} & \textbf{78.5} & \textbf{De} & \textbf{23.8} & \textbf{82.1} & De    & 29.8  & 84.0 \\

    \midrule
    
    \multirow{6}{*}{\textbf{Bloomz-176B}} & \textbf{Es} & \textbf{25.0} & \textbf{82.8} & \textbf{Es} & \textbf{20.8} & \textbf{80.9} & \textbf{En} & \textbf{34.6} & \textbf{82.1} & Es    & 6.1   & 63.6  & Es    & 27.3  & 82.8  & \textbf{Es} & \textbf{31.5} & \textbf{84.6} \\
          & Ru    & 17.5  & 76.0  & Ru    & 14.8  & 75.2  & Ru    & 22.2  & 75.1  & \textbf{Ru} & \textbf{9.5} & \textbf{66.2} & Ru    & 22.2  & 79.1  & Ru    & 20.4  & 77.5 \\
          & Fr    & 24.9  & 82.6  & Fr    & 19.7  & 80.2  & Es    & 33.5  & 81.5  & Fr    & 8.9   & 67.1  & \textbf{Fr} & \textbf{27.6} & \textbf{82.6} & Fr    & 29.9  & 84.3 \\
          & Zh    & 17.1  & 79.2  & Ja    & 13.2  & 74.5  & Zh    & 21.0  & 78.0  & Zh    & 7.3   & 66.3  & Ja    & 17.2  & 78.9  & It    & 28.9  & 82.4 \\
          & Ja    & 13.0  & 74.3  & Cs    & 10.7  & 66.4  & It    & 32.2  & 80.3  & Ja    & 4.9   & 60.9  & Cs    & 15.1  & 68.8  & Cs    & 14.5  & 67.8 \\
          & Cs    & 13.6  & 64.7  & De    & 17.3  & 77.7  & Cs    & 15.1  & 64.0  & Cs    & 2.5   & 51.9  & De    & 25.5  & 79.6  & De    & 26.8  & 81.5 \\

    \bottomrule
    
    \end{tabular}
    
}

  \caption{Full experimental results of pivot prompts on WMT dataset. The best results of each group are in \textbf{bold}.}
  \label{app tab: Full experimental results of pivot prompts on WMT dataset}
\end{table*}

\begin{table*}[htbp]
\LARGE
  \centering
  
 \resizebox{0.7\linewidth}{!}{
 
    \begin{tabular}{l|cc|cc|cc|cc|cc}
    \toprule
    \multirow{3.5}{*}{\textbf{Model}} & \multicolumn{2}{c|}{\textbf{RTE}} & \multicolumn{6}{c|}{\textbf{XNLI}}  & \multicolumn{2}{c}{\textbf{BoolQ}} \\
    \cmidrule(lr){2-3} \cmidrule(lr){4-9} \cmidrule(lr){10-11}
    & \multicolumn{2}{c|}{\textit{En}} & \multicolumn{2}{c|}{\textit{Fr}} & \multicolumn{2}{c|}{\textit{De}} & \multicolumn{2}{c|}{\textit{Zh}} & \multicolumn{2}{c}{\textit{En}} \\
    \cmidrule(lr){2-3} \cmidrule(lr){4-9} \cmidrule(lr){10-11}
    & Pivot & Accuracy & Pivot & Accuracy & Pivot & Accuracy & Pivot & Accuracy & Pivot & Accuracy \\

    \midrule    
    \multirow{3}{*}{\textbf{Qwen-7B}} & De & 85.9 & \textbf{De} & \textbf{78.9} & \textbf{Es} & \textbf{80.2} & \textbf{De} & \textbf{74.2} & \textbf{Es} & \textbf{81.6} \\
    & \textbf{Es} & \textbf{86.6} & Es & 77.9 & Fr & 79.2 & Es & 74.1 & - & - \\
    & Fr & 85.6 & Ru & 77.2 & Ru & 77.2 & Fr & 72.3 & - & - \\

    \midrule    
    \multirow{3}{*}{\textbf{Qwen-14B}} & De & 89.2 & De & 80.1 & Es & 79.5 & De & 73.3 & \textbf{Es} & \textbf{86.0}  \\
    & \textbf{Es} & \textbf{90.6} & \textbf{Es} & \textbf{80.5} & \textbf{Fr} & \textbf{79.8} & \textbf{Es} & \textbf{74.2} & - & - \\
    & Fr & 88.8 & Ru & 79.1 & Ru & 77.7 & Fr & 72.8 & - & - \\

    \midrule    
    \multirow{3}{*}{\textbf{ALMA-13B}} & De & 84.1 & \textbf{De} & \textbf{82.0} & Es & 79.6 & \textbf{De} & \textbf{75.9} & \textbf{Es} & \textbf{77.7} \\
    & \textbf{Es} & \textbf{84.5} & Es & 81.7 & \textbf{Fr} & \textbf{80.8} & Es & 74.3 & - & - \\
    & Fr & 80.1 & Ru & 79.4 & Ru & 79.8 & Fr & 74.6 & - & - \\

    \midrule    
    \multirow{3}{*}{\textbf{Yi-34B}} & De & 79.1 & De & 70.0 & \textbf{Es} & \textbf{72.6} & De & 64.7 & \textbf{Es} & \textbf{84.2} \\
    & \textbf{Es} & \textbf{85.9} & \textbf{Es} & \textbf{71.5} & Fr & 71.9 & \textbf{Es} & \textbf{68.1} & - & - \\
    & Fr & 84.8 & Ru & 66.6 & Ru & 64.8 & Fr & 66.6 & - & - \\

    \midrule    
    \multirow{3}{*}{\textbf{Qwen-72B}} & De & 91.3 & \textbf{De} & \textbf{85.8} & \textbf{Es} & \textbf{85.5} & De & 78.9 & \textbf{Es} & \textbf{88.7} \\
    & \textbf{Es} & \textbf{92.4} & Es & 85.0 & Fr & 85.2 & \textbf{Es} & \textbf{80.6} & - & - \\
    & Fr & 90.6 & Ru & 83.9 & Ru & 83.5 & Fr & 79.5 & - & - \\

    \midrule    
    \multirow{3}{*}{\textbf{Bloomz-176B}} & De & 74.4 & De & 50.0 & Es & 53.0 & De & 49.6 & - & - \\
    & Es & 73.3 & \textbf{Es} & \textbf{53.1} & Fr & 50.5 & \textbf{Es} & \textbf{53.7} & - & - \\
    & \textbf{Fr} & \textbf{77.6} & Ru & 50.8 & \textbf{Ru} & \textbf{53.3} & Fr & 52.0 & - & - \\
    
    \bottomrule
    \end{tabular}    
}

  \caption{Full experimental results of pivot prompts on RTE, XNLI and BoolQ dataset. The best results of each group are in \textbf{bold}.}
  \label{app tab: Full experimental results of pivot prompts on RTE, XNLI and BoolQ dataset}
\end{table*}

\onecolumn
{\small
\begin{xltabular}{0.8\textwidth}{ccX}

    \toprule
    \textbf{Dataset} & \multicolumn{2}{c}{\textbf{Prompt}}\\ 
    \midrule
    \endhead
    
    \multicolumn{3}{r}{{Continued on next page}} \\ 
    \bottomrule \\
    \endfoot
    
    \bottomrule \\
    \caption{All the prompts used in experiments.}  \label{app tab: All the prompts used in experiments} \\
    \endlastfoot

    \multirow{25}{*}{\makecell[c]{\textbf{FLORES-200} \\ \\ \textbf{WMT}}} & Direct & \makecell[X]{Translate into \colorbox{gray!30}{\textit{target-language}}.\\  \colorbox{gray!30}{\textit{source-language}}: \colorbox{gray!30}{\textit{source-sentence}}\\  \colorbox{gray!30}{\textit{target-language}}: } \\ 
    \cmidrule(lr){2-3}
    & \textsc{PMI} & \makecell[X]{Translate into \colorbox{gray!30}{\textit{target-language}}.\\ \colorbox{gray!30}{\textit{source-language}}: \colorbox{gray!30}{\textit{source-sentence}}\\ \colorbox{gray!30}{\textit{parallel-language(1)}}: \colorbox{gray!30}{\textit{parallel-sentence(1)}}\\ \colorbox{gray!30}{\textit{parallel-language(2)}}: \colorbox{gray!30}{\textit{parallel-sentence(2)}}\\  ······\\ \colorbox{gray!30}{\textit{parallel-language(n)}}: \colorbox{gray!30}{\textit{parallel-sentence(n)}}\\ \colorbox{gray!30}{\textit{target-language}}:}  \\ 
    \cmidrule(lr){2-3}
    & \makecell[c]{\textsc{PMI}$_{MS}$ \\ \textsc{PMI}$_{PA}$} & \makecell[X]{There are six sentences in \colorbox{gray!30}{\textit{source-language}}, I need you to fully understand all of them and then translate to one \colorbox{gray!30}{\textit{target-language}} sentence.\\ \colorbox{gray!30}{\textit{source-language}}:\\ 1. \colorbox{gray!30}{\textit{paraphrase-sentence1}}\\ 2. \colorbox{gray!30}{\textit{paraphrase-sentence2}}\\ 3. \colorbox{gray!30}{\textit{paraphrase-sentence3}}\\ 4. \colorbox{gray!30}{\textit{paraphrase-sentence4}}\\ 5. \colorbox{gray!30}{\textit{paraphrase-sentence5}}\\ \colorbox{gray!30}{\textit{target-language}}:}  \\

    \midrule
    \multirow{20}{*}{\makecell[c]{\textbf{WikiAuto}}} & Direct & \makecell[X]{You will be presented with a complex sentence. Your task is to simplify this sentence to make it easier to understand, while maintaining its core meaning and factual content. The goal is to generate a simplified version of the sentence without losing important information or altering its original intent. Please provide a single simplified sentence as your response, without any explanation. Here is the complex sentence:\\ Complex Sentence: \colorbox{gray!30}{\textit{sentence}}\\ Your simplified version:} \\ 
    \cmidrule(lr){2-3}
    & \textsc{PMI} & \makecell[X]{You will be presented with the same sentence in four different languages: \colorbox{gray!30}{\textit{source-language}}, \colorbox{gray!30}{\textit{parallel-language1}}, \colorbox{gray!30}{\textit{parallel-language2}}, and \colorbox{gray!30}{\textit{parallel-language3}}. These sentences convey the exact same meaning. Your task is to simplify the sentence into \colorbox{gray!30}{\textit{source-language}} to make it easier to understand, while maintaining its core meaning and factual content. It is important to note that since all sentences have the same meaning, you only need to provide one simplified \colorbox{gray!30}{\textit{source-language}} version. Please generate a single simplified \colorbox{gray!30}{\textit{source-language}} sentence as your response, without any explanation. Here are the sentences:\\ \colorbox{gray!30}{\textit{source-language}} Sentence: \colorbox{gray!30}{\textit{source-sentence}}\\ \colorbox{gray!30}{\textit{parallel-language1}} Sentence: \colorbox{gray!30}{\textit{parallel-sentence1}}\\ \colorbox{gray!30}{\textit{parallel-language2}} Sentence: \colorbox{gray!30}{\textit{parallel-sentence2}}\\ \colorbox{gray!30}{\textit{parallel-language3}} Sentence: \colorbox{gray!30}{\textit{parallel-sentence3}}\\ Your simplified \colorbox{gray!30}{\textit{source-language}} version:}  \\ 

    \midrule
    \multirow{33}{*}{\textbf{RTE}} & Direct & \makecell[X]{You will be presented with a pair of sentences.Your task is to determine the relationship between these two sentences. There are two possible relationships: entailment, not\_entailment. 'entailment' means the first sentence logically implies the second one. 'not\_entailment' means the first sentence logically conflicts with the second one. Please provide a single prediction for the relationship based on these sentence pairs, without any explanation. Here is the sentence pair:\\Premise: \colorbox{gray!30}{\textit{src-premise}}\\Hypothesis: \colorbox{gray!30}{\textit{src-hypothesis}}\\Your prediction:}  \\ 
    \cmidrule(lr){2-3}
    & \makecell[c]{\textsc{PMI}} & \makecell[X]{You will be provided with a set of sentence pairs that are semantically identical but presented in four different languages: \colorbox{gray!30}{\textit{src-language}}, \colorbox{gray!30}{\textit{parallel-language1}}, \colorbox{gray!30}{\textit{parallel-language2}}, and \colorbox{gray!30}{\textit{parallel-language3}}. Each pair consists of a premise and a hypothesis. Despite the language differences, the meaning of these sentences is the same across all languages. Your task is to analyze these sentence pairs and determine the relationship between the premise and the hypothesis. There are two possible relationships: entailment and not\_entailment. 'entailment' means the first sentence logically implies the second one. 'not\_entailment' means the first sentence logically conflicts with the second one. Please provide a single prediction for the relationship based on these sentence pairs, without any explanation. Here are the sentence pairs:\\ \colorbox{gray!30}{\textit{src-language}}:\\ Premise: \colorbox{gray!30}{\textit{src-premise}}\\ Hypothesis: \colorbox{gray!30}{\textit{src-hypothesis}}\\ \colorbox{gray!30}{\textit{parallel-language1}}:\\ Premise: \colorbox{gray!30}{\textit{para1-premise}}\\ Hypothesis: \colorbox{gray!30}{\textit{para1-hypothesis}}\\ \colorbox{gray!30}{\textit{parallel-lang2}}:\\ Premise: \colorbox{gray!30}{\textit{para2-premise}}\\ Hypothesis: \colorbox{gray!30}{\textit{para2-hypothesis}}\\ \colorbox{gray!30}{\textit{parallel-lang3}}:\\ Premise: \colorbox{gray!30}{\textit{para3-premise}}\\ Hypothesis: \colorbox{gray!30}{\textit{para3-hypothesis}}\\ Your prediction:}  \\

    \midrule
    \multirow{14}{*}{\textbf{XLSum}} & Direct & \makecell[X]{You will be presented with a long text. Your task is to summarize this text in 1-2 sentences in \colorbox{gray!30}{\textit{source-language}}, capturing the most important and core content. The summary should distill the essence of the article concisely and accurately. Please provide a single summary for the text without any explanation. Here is the text:\\ \colorbox{gray!30}{\textit{source-text}}\\ Your summary:} \\ 
    \cmidrule(lr){2-3}
    & \textsc{PMI} & \makecell[X]{You will be presented with two texts, each in a different language: \colorbox{gray!30}{\textit{source-language}}, \colorbox{gray!30}{\textit{parallel-language}}. These texts convey the same meaning in their respective languages. Your task is to summarize the core content of these texts in one summary (1-2 sentences) in \colorbox{gray!30}{\textit{source-language}}, capturing the most important and central idea. Please provide a single summary for the texts without any explanation. Here are the texts:\\\colorbox{gray!30}{\textit{source-language}} Text: \colorbox{gray!30}{\textit{source-text}}\\\colorbox{gray!30}{\textit{parallel-language}} Text: \colorbox{gray!30}{\textit{parallel-text}}\\Your summary in \colorbox{gray!30}{\textit{source-language}}:}  \\

    \midrule
    \multirow{14}{*}{\textbf{BoolQ}} & Direct & \makecell[X]{You will be provided with a passage and a yes/no question based on that passage. Your task is to read the passage and then answer the question with a simple `Yes' or `No' based on the information in the passage. Please do not provide any explanations or reasoning for your answer.\\ Passage: \colorbox{gray!30}{\textit{source-passage}}\\ Question: \colorbox{gray!30}{\textit{source-question}}\\ Please respond with `Yes' or `No' only. Your answer:}  \\ 
    \cmidrule(lr){2-3}
    & \textsc{PMI} & \makecell[X]{You will be provided with two passages, each in a different language: \colorbox{gray!30}{\textit{source-language}}, \colorbox{gray!30}{\textit{parallel-language}}. These passages convey the same meaning. Your task is to understand the content of these passages and then answer a yes/no question based on them. It's important to note that you only need to make one prediction as the semantic content across all the passages is identical. Please do not provide any explanations or reasoning for your answer.\\ \colorbox{gray!30}{\textit{source-language}} Passage: \colorbox{gray!30}{\textit{source-sentence}}\\ \colorbox{gray!30}{\textit{parallel-language}} Passage: \colorbox{gray!30}{\textit{parallel-sentence}}\\ Question: \colorbox{gray!30}{\textit{source-question}}\\ Please respond with `Yes' or `No' only. Your answer:}  \\

    \midrule
    \multirow{20}{*}{\textbf{XNLI}} & Direct & \makecell[X]{You will be presented with a pair of sentences. Your task is to determine the relationship between these two sentences. There are three possible relationships: entailment, contradiction, or neutral. Please provide a single prediction for the relationship based on these sentence pairs, without any explanation. Here is the sentence pair:\\  Premise: \colorbox{gray!30}{\textit{premise-sentence}}\\  Hypothesis: \colorbox{gray!30}{\textit{hypothesis-sentence}}\\  Your prediction:}  \\
    \cmidrule(lr){2-3}
    & \textsc{PMI} & \makecell[X]{You will be given a premise in multiple languages (\colorbox{gray!30}{\textit{source-language}}, \colorbox{gray!30}{\textit{parallel-language1}}, \colorbox{gray!30}{\textit{parallel-language2}}, \colorbox{gray!30}{\textit{parallel-language3}}) and a hypothesis in \colorbox{gray!30}{\textit{source-language}}. Your task is to determine the relationship between the multilingual premises and the \colorbox{gray!30}{\textit{source-language}} hypothesis. There are three possible relationships: entailment, contradiction, or neutral. Please provide a single prediction for the relationship, without any explanation. Here are the premises and the hypothesis:\\  \colorbox{gray!30}{\textit{source-sentence}} Premise: \colorbox{gray!30}{\textit{source-premise}}\\  \colorbox{gray!30}{\textit{parallel-language1}} Premise: \colorbox{gray!30}{\textit{parallel-premise1}}\\  \colorbox{gray!30}{\textit{parallel-language2}} Premise: \colorbox{gray!30}{\textit{parallel-premise2}}\\  \colorbox{gray!30}{\textit{parallel-language3}} Premise: \colorbox{gray!30}{\textit{parallel-premise3}}\\  Hypothesis: \colorbox{gray!30}{\textit{source-hypothesis}}\\  Your prediction:}  \\

    \midrule
    \multirow{14}{*}{\textbf{GSM8K}} & Direct & \makecell[X]{Q: \colorbox{gray!30}{\textit{source-sentence}} \\  A:}  \\
    \cmidrule(lr){2-3}
    & \textsc{PMI} & \makecell[X]{You are provided with a set of parallel mathematical problems in multiple languages. Each problem presents the same mathematical question, but expressed in different languages. Your task is to comprehend the problem in any of these languages, reason through the problem in English, and finally, generate a solution in English.\\  Question in English: \colorbox{gray!30}{\textit{source-sentence}} \\  Question in \colorbox{gray!30}{\textit{parallel-language}}: \colorbox{gray!30}{\textit{parallel-sentence}} \\ Question in \colorbox{gray!30}{\textit{parallel-language}}: \colorbox{gray!30}{\textit{parallel-sentence}} \\ Question in \colorbox{gray!30}{\textit{parallel-language}}: \colorbox{gray!30}{\textit{parallel-sentence}} \\  Answer in English:}  \\

\end{xltabular}
}
\twocolumn

\end{document}